\documentclass[12pt]{article} 

\usepackage[dvips]{graphics}
\DeclareGraphicsExtensions{.eps.gz,.eps,.epsi.gz,.epsi,.ps,.ps.gz}
\usepackage{epsfig}
\usepackage{color}
\graphicspath{{fig/}} \textwidth=6.5in \textheight=9in

\usepackage{latexsym}
\usepackage{graphicx}
\usepackage{amsmath}
\usepackage{amsfonts}
\usepackage{amssymb}
\usepackage{mathrsfs}
\usepackage{verbatim}
\usepackage{geometry}

\def\argmin{\mathop{\rm arg\, min}}

\newcommand{\bel}{\begin{eqnarray}\label}
\newcommand{\eel}{\end{eqnarray}}
\newcommand{\bes}{\begin{eqnarray*}}
\newcommand{\ees}{\end{eqnarray*}}

\def\benu{\begin{enumerate}}
\def\eenu{\end{enumerate}}

\def\argmin{\mathop{\rm arg\, min}}
\def\real{{\mathbb{R}}}

\def\complex{\mathop{{\rm I}\kern-.58em\hbox{\rm C}}\nolimits}
\def\pa{\partial}

\def\diag{\hbox{\rm diag}}

\def\sgn{\hbox{\rm sgn}}
\def\trace{\hbox{\rm trace}}

\def\scrB{{\mathscr B}}

\def\scrC{{\mathscr C}}

\def\scrC{{\mathscr C}}

\def\what{\widehat{w}}

\def\What{\widehat{W}}

\def\ytil{\widetilde{y}}

\def\ztil{\widetilde{z}}

\def\hbeta{\widehat{\beta}}
\def\tbeta{\widetilde{\beta}}

\def\eps{\epsilon}\def\veps{\varepsilon}

\def\lam{\lambda}

\def\hsigma{\widehat{\sigma}}

\def\Sigmahat{\widehat{\Sigma}}

\newtheorem{theorem}{Theorem}
\newtheorem{lemma}{Lemma}

\newtheorem{corollary}{Corollary}
\newtheorem{remark}{Remark}[section]
\newtheorem{example}{Example}[section]

\def\argmin{\mathop{\rm arg\, min}}
\def\real{\mathop{{\rm I}\kern-.2em\hbox{\rm R}}\nolimits}

\def\dell{{\dot\ell}}
\def\ddell{{\ddot\ell}}
\def\drho{{\dot\rho}}
\def\dpsi{{\dot\psi}}
\def\ddpsi{{\ddot\psi}}

\def\rE{{\mathrm E}}

\def\rP{{\mathrm P}}

\def\hw{{\what}}
\def\hW{{\What}}

\def\tbeta{\tilde{\beta}}

\begin{document}

\centerline{\bf \large Estimation And Selection Via}
\centerline{\bf \large Absolute Penalized Convex Minimization}
\centerline{\bf \large And Its Multistage Adaptive Applications}
\bigskip

\centerline{Jian Huang and Cun-Hui Zhang\footnote
{Jian Huang's research is partially supported  by NIH Grants CA120988, CA142774 and NSF Grant DMS 0805670.
Cun-Hui Zhang's research is partially supported by NSF Grants DMS 0604571, DMS 0804626
and NSA Grant H98230-09-1-0006.}}
\bigskip

\centerline{University of Iowa and Rutgers University}


\bigskip

\noindent

\noindent
{\it Abstract:}
The $\ell_1$-penalized method, or the Lasso,
has emerged as an important tool for the analysis of large data sets.
Many important results have been obtained for the Lasso in linear regression
which have led to a deeper understanding of high-dimensional statistical problems.
In this article, we consider a class of weighted $\ell_1$-penalized
estimators for convex loss functions of a general form, including the generalized
linear models.
We study  the estimation, prediction, selection and sparsity  properties
of the weighted $\ell_1$-penalized estimator in sparse, high-dimensional settings
where the number of predictors $p$ can be much larger than the sample size $n$.
Adaptive Lasso is considered as a special case.  A multistage method is developed
to apply an adaptive Lasso recursively.
We provide $\ell_q$ oracle inequalities, a general selection consistency theorem,
and an upper bound on the dimension of the Lasso estimator.
Important models including
the linear regression, logistic regression and log-linear models are
used throughout to illustrate the applications of the general results.

\medskip
\noindent
{\it Running title:} Absolute penalized convex minimization

\noindent
{\it Key words:} 
Variable selection, penalized estimation,
oracle inequality, generalized linear models, selection consistency,
sparsity.

\section{Introduction}

High-dimensional data arise in many diverse fields of scientific research.
For example, in genetic and genomic studies, more and more large data sets are being generated with rapid advances in biotechnology, where the total number of variables $p$ is larger than the sample size $n$. Fortunately, statistical analysis is still
possible for a substantial subset of such problems with a sparse underlying model
where the number of important variables is
much smaller than the sample size.
A fundamental problem in the analysis of such data is to find reasonably accurate sparse
solutions that are easy to interpret and can be
used for the prediction and estimation of covariable effects.
The $\ell_1$-penalized method, or the Lasso \cite{Tibshirani96, ChenDS98},
has emerged as an important approach to finding such solutions in sparse, high-dimensional statistical problems.

In the last few years, considerable progress has been made in
understanding the theoretical properties of the Lasso in
 $p \gg n $ settings.  Most results have been obtained
for linear regression models with a quadratic loss.
\cite{GreenshteinR04} studied the prediction performance of the Lasso
in high-dimensional least squares regression.
\cite{MeinshausenB06} showed
that, for neighborhood selection in the Gaussian graphical models,
under a neighborhood stability condition on the design matrix and
certain additional regularity conditions, the Lasso is selection consistent
even when $p \rightarrow \infty$ at a rate faster than $n$.
\cite{ZhaoY06} formalized the neighborhood
stability condition in the context of linear regression as
a strong irrepresentable condition.
\cite{CandesT07} derived an upper bound for the $\ell_2$ loss for the
estimation of regression coefficients with a closely related Dantzig selector
under a condition on the number of nonzero coefficients and a uniform uncertainty
principle on the design matrix.
Similar results have been obtained for the Lasso.
For example, upper bounds for the $\ell_q$ loss of the Lasso estimator has being
established by \cite{BuneaTW07} for $q = 1$,
\cite{ZhangH08} for $q \in  [1; 2]$,
\cite{MeinshausenY09} for $q = 2$, \cite{BickelRT09} for $q \in [1; 2]$,
and \cite{Zhang09-l1,YeZ10}
for general $q \ge 1$.
For convex minimization methods beyond linear regression, \cite{vandeGeer08}
studied the Lasso in high-dimensional generalized linear models
(GLM) and obtained prediction and $\ell_1$ estimation error bounds. \cite{NegahbanRWY10}
studied  penalized M-estimators
with a general class of regularizers, including an $\ell_2$ error
bound for the Lasso in GLM under a restricted convexity and other regularity conditions.

Theoretical studies of the Lasso have revealed that it may not perform well for the
purpose of variable selection, since its required irrepresentable condition
is not properly scaled in the number of relevant variables. In a number of
simulation studies, the Lasso has shown weakness in variable selection when
the number of nonzero regression coefficients increases.
As a remedy, a number of proposals have been introduced in the literature,
including concave penalized LSE \cite{FanL01, Zhang10-mc+},  adaptive Lasso \cite{Zou06},
and stepwise regression \cite{Zhang11-foba}.
Although extensions of the concave penalized
LSE is beyond the scope of this paper, adaptive Lasso is studied here as
a weighted Lasso with estimated weights.
When the number of predictors $p$ is fixed, \cite{Zou06}
proved that the adaptive Lasso has the asymptotic oracle property.
In linear regression models.
\cite{HuangMZ08}
showed that the oracle property continues to hold for the adaptive Lasso in $p \gg n$ settings
under an adaptive irrepresentable and other regularity conditions.
\cite{MeierB07} suggested using the Lasso as the initial estimator
for the adaptive Lasso or even a multi-step adaptive Lasso.
The one-step method of \cite{ZouL08}, designed to approximate penalized
estimators with concave penalties, can be also viewed as adaptive Lasso.

In this article, we consider a class of weighted $\ell_1$-penalized
estimators with a  convex loss function. This class includes the Lasso, adaptive Lasso
and multistage recursive application of an adaptive Lasso
in generalized linear models as special cases.
We study  the estimation, prediction, selection and
sparsity  properties of the weighted $\ell_1$-penalized estimator based
on a convex loss in sparse, high-dimensional settings where the number of predictors $p$
can be much larger than the sample size $n$.
The main contributions of this work are as follows.
\begin{itemize}
\item We extend the existing theory for the unweighted Lasso from linear regression to more
general convex loss function.

\item We develop a multistage method with recursive applications of an adaptive Lasso
and provide sharper risk bound than those for unweighted Lasso.

\item We apply our general results to a number of  
important special cases, including the linear, logistic and log-linear regression models.

\end{itemize}

This article is organized as follows. In Section \ref{apcm} we describe a general formulation of the absolute penalized minimization problem with a
convex loss, along with two basic inequalities and a number of examples.
In Section \ref{oracle-ineq} we develop oracle inequalities
for the weighted Lasso estimator for general quasi star-shaped loss functions
and an $\ell_2$ bound on the prediction error.
In Section \ref{multistage-section} we develop sharper oracle inequalities for multistage recursive applications of an adaptive Lasso.
In Section \ref{selection-consistency} we derive sufficient conditions for
selection consistency.
In Section \ref{sparsity} we provide an upper bound on the dimension
of the Lasso estimator.
Concluding remarks are given in Section \ref{discussion}.
All proofs are provided in an appendix.

\section{Absolute penalized convex minimization}
\label{apcm}
\subsection{Definition and the KKT conditions}
We consider a general convex loss function of the form
\bel{model}
\ell(\beta) = \psi(\beta) - \langle \beta, z \rangle,
\eel
where $\psi(\beta)$ is a known convex function, $z$ is observed and $\beta$ is unknown.
Unless otherwise stated, the inner product space is $\real^p$, so that
$\{z,\beta\}\subset\real^p$ and $\langle \beta, z\rangle = \beta'z$.
Our analysis of (\ref{model}) requires certain smoothness of the function $\psi(\beta)$
in terms of its differentiability. In what follows, such smoothness assumptions
are always explicitly described by invoking the derivative of $\psi$.
For any $v =(v_1, \ldots, v_p)'$,
we use $\|v\|$ to denote a general norm of $v$ and $|v|_q$ the $\ell_q$ norm
$(\sum_j |v_j|^q)^{1/q}$, with $|v|_{\infty} =\max_j|v_j|$.
Let $\hw\in\real^p$ be a (possibly estimated) weight vector
with nonnegative elements $\hw_j, 1 \le j \le p$,
and $\hW=\diag(\hw)$.
The weighted absolute penalized estimator, or weighted Lasso, is defined as
\bel{Lasso}
\hbeta = \argmin_{\beta} \Big\{\ell(\beta) + \lam|\hW\beta|_1\Big\}.
\eel

Here we focus on the case where $\hW$ is diagonal. In linear regression, 
\cite{TibshiraniT11} considered non-diagonal, predetermined $\What$ and
derived an algorithm for computing the solution paths.

A vector $\hbeta$ is a global minimizer in (\ref{Lasso})
if and only if the negative gradient at $\hbeta$ satisfies the Karush-Kuhn-Tucker (KKT) conditions,
\bel{KKT}
g = -\dell(\hbeta) = z - \dpsi(\hbeta),\
\begin{cases}
g_j = \hw_j\lam\,\sgn(\hbeta_j) & \hbox{ if }\hbeta_j\neq 0 \cr
g_j \in \hw_j\lam [-1,1] & \hbox{ all }j,
\end{cases}
\eel
where $\dell(\beta) = (\pa/\pa\beta)\ell(\beta)$ and $\dpsi(\beta) = (\pa/\pa\beta)\psi(\beta)$.
Since the KKT conditions are necessary and sufficient for (\ref{Lasso}),
results on the performance of $\hbeta$ can be viewed as analytical
consequences of (\ref{KKT}).

The estimator (\ref{Lasso}) includes the $\ell_1$-penalized estimator, or the Lasso,
with the choice $\hw_j=1, 1 \le j \le p$. 
A careful study of the (unweighted) Lasso
in general convex minimization (\ref{model}) is by itself an interesting and important problem.
Our work includes the Lasso as a special case since $\hw_j=1$ is allowed in all our theorems.

In practice, unequal $\what_j$ arise in many ways.
In adaptive Lasso \cite{Zou06}, a decreasing function of a certain initial estimator of $\beta_j$ is used as the weight $\hw_j$ to remove the bias of the Lasso.
In \cite{FanL01, ZouL08, Zhang10-multistage}, the weights $\what_j$ are computed iteratively
with $\what_j = \drho_\lam(\hbeta_j)$, where $\drho_\lam(t)=(d/dt)\rho_\lam(t)$ with
a suitable concave penalty function $\rho_\lam(t)$. This is also designed to remove the bias of the Lasso, since the concavity of $\rho_\lam(t)$ guarantees smaller weight for larger $\hbeta_j$.
In Section 4, we provide results on the improvements
of this weighted Lasso over the standard Lasso. 
In linear regression, \cite{Zhang10-multistage} gave suitable conditions under which this iterative algorithm provides smaller weights $\what_j$ for most large $\beta_j$. Such nearly unbiased methods are expected to
produce better results than the Lasso when a significant fraction of nonzero $|\beta_j|$ are of the order $\lam$ or larger. 
Regardless of the computational methods,
the results in this paper demonstrate 
the benefits of using data dependent weights in a general class of
problems with convex losses. 

Unequal weights may also arise for computational reasons.
The Lasso with $\what_j=1$ is expected to perform similarly to weighted Lasso with
data dependent $1\le \what_j\le C_0$, with a fixed $C_0$. However, the weighted
Lasso is easier to compute since $\what_j$ can be determined as a part of
an iterative algorithm.
For example, in a gradient descent algorithm, one may take larger steps and stop the computation as
soon as the KKT conditions (\ref{KKT}) are attained for any weights satisfying $1\le \what_j\le C_0$.

The weight function $\what_j$ can be also used to standardize the penalty level,
for example with $\what_j = \{\ddpsi_{jj}(\hbeta)\}^{1/2}$, where $\ddpsi_{jj}(\beta)$
is the $j$-th diagonal element of the Hessian matrix of $\psi(\beta)$.
When $\psi(\beta)$ is quadratic, for example in linear regression, $\what_j$ does
not depend on $\hbeta$. However, in other convex minimization problems,
such weights need to be computed iteratively.

Finally, in certain applications, the effects of a certain set $S_*$ of variables are of primary interest, so that penalization of $\hbeta_{S_*}$, and thus the resulting bias, should be avoided. This leads to ``semi-penalized'' estimators with $\hw_j=0$ for $j\in S_*$, for example, with $\hw_i = I\{j\not\in S_*\}$.

\subsection{Basic inequalities, prediction, and Bregman divergence}

Let $\beta^*$ denote a target vector for $\beta$.
In high-dimensional models, the performance of an estimator $\hbeta$ is typically measured
by its proximity to a target under conditions on the sparsity of $\beta^*$ and the size of the
negative gradient $ - \dell(\beta^*) = z - \dpsi(\beta^*)$.
For $\ell_1$-penalized estimators, such results are often derived from
the KKT conditions (\ref{KKT}) via certain basic inequalities, which are direct consequences of the KKT conditions and have appeared in different forms in the literature, for example, in the papers cited in the Introduction.
Let $D(\beta,\beta^*)=\ell(\beta)-\ell(\beta^*)-\langle \dell(\beta^*),\beta-\beta^*\rangle$ be
the Bregman divergence \cite{Bregman67} and consider its symmetrized version \cite{NielsenN07}
\bel{Delta}
\Delta(\beta,\beta^*)
= D(\beta,\beta^*) + D(\beta^*,\beta)
= \big\langle \beta-\beta^*, \dpsi(\beta)-\dpsi(\beta^*)\big\rangle.
\eel
Since $\psi$ is convex, $\Delta(\beta,\beta^*)\ge 0$.
Two basic inequalities below provide upper bounds for the symmetrized Bregman
divergence $\Delta(\hbeta,\beta^*)$.
The sparsity of $\beta^*$ is measured by a weighted $\ell_1$ norm of $\beta^*$
in the first one and by the number of zero entries in the second one.

Let $S$ be any set of indices satisfying
$S \supseteq \{j: \beta^*_j\neq 0\}$ and let $S^c$ be the complement of $S$ in $\{1, \ldots, p\}$.
We shall refer to $S$ as the sparse set.
Let $W=\diag(w)$ for a possibly unknown vector $w\in\real^p$ with elements $w_j\ge 0$.
Define
\bel{z^*}
&& z^*_0 = |\{z - \dpsi(\beta^*)\}_S|_\infty,
\ z^*_1 = |W_{S^c}^{-1}\{z - \dpsi(\beta^*)\}_{S^c}|_\infty,
\\ \label{Omega_0} && \Omega_0 = \big\{\what_j \le w_j\ \forall j\in S\big\}
\cap\big\{ w_j\le \what_j\ \forall j\in S^c\big\},
\eel
where for any $p$-vector $v$ and set $A$, $v_{A}=(v_j: j \in A)'$.
Here and in the sequel $M_{AB}$ denotes the $A\times B$ subblock of a matrix
$M$ and $M_A=M_{AA}$.

\begin{lemma}
\label{Lem1} (i) Let $\beta^*$ be a target vector.
In the event $\Omega_0\cap\{|(z-\dpsi(\beta^*))_j|\le \hw_j\lam\ \forall j\}$,
\bel{pred}
\Delta(\hbeta,\beta^*) \le 2\lam |\hW\beta^*|_1\le 2\lam |W\beta^*|_1.
\eel
(ii) For any target vector $\beta^*$ and $S\supseteq \{j:\beta^*_j \neq 0\}$, the error
$h=\hbeta-\beta^*$ satisfies
\bel{basic-ineq}
\Delta(\beta^*+h,\beta^*) + (\lam - z^*_1)|W_{S^c}h_{S^c}|_1
&\le& \langle h_S, g_S - \{z-\dpsi(\beta^*)\}_S \rangle
\cr &\le& (|w_S|_\infty\lam + z^*_0)|h_S|_1
\eel
in $\Omega_0$ for a certain negative gradient vector $g$ satisfying $|g_j|\le \what_j\lam$.
Consequently, in $\Omega_0\cap\{(|w_S|_\infty\lam + z^*_0)/(\lam - z^*_1)\le\xi\}$,
$h\neq 0$ belongs to the sign-restricted cone
$\scrC_-(\xi,S) =\{b\in\scrC(\xi,S): b_j(\dpsi(\beta+b)-\dpsi(\beta))_j \le 0\ \forall j\in S^c\}$,
where
\bel{cone}
\scrC(\xi,S) = \big\{b\in\real^p: |W_{S^c}b_{S^c}|_1\le \xi |b_S|_1\neq 0\big\}.
\eel
\end{lemma}

\begin{remark}
Sufficient conditions are given in Subsection 3.2 for
$\{|(z-\dpsi(\beta^*))_j|\le \hw_j\lam\ \forall j\}$
to hold with high probability in generalized linear models. See Lemma \ref{Lem2},
Remarks \ref{Remark3p1a} and \ref{Remark3p1} and Examples
\ref{example-linearB}, \ref{example-logisticB}, and \ref{example-loglinearB}.
\end{remark}

\medskip
A useful feature of Lemma \ref{Lem1} is the explicit statements of
the monotonicity of the basic inequality in the weights.
By Lemma \ref{Lem1} (ii), it suffices to study the analytical properties of the penalized
criterion with the error $h=\hbeta-\beta^*$ in the sign-restricted cone,
provided that the event $(|w_S|_\infty\lam + z^*_0)/(\lam - z^*_1)\le\xi$
has large probability.
However, unless $\scrC_-(\xi,S)$ is specified, we will consider the larger cone in
(\ref{cone}) in order to simplify the analysis.
The choices of the target vector $\beta^*$, the sparse set $S\supseteq\{j:\beta^*_j\neq 0\}$,
weight vector $\hw$ and its bound $w$
are quite flexible. The main requirement is that $\{|S|,z^*_0,z^*_1\}$ should be small.
In linear regression or generalized linear models, we may conveniently consider
$\beta^*$ as the vector of true regression coefficients under a probability measure $\rP_{\beta^*}$.
However, $\beta^*$ can also be a sparse version of a  true $\beta$, e.g.
$\beta^*_j = \beta_jI\{|\beta_j| \ge \tau\}$ for a threshold value $\tau$ under $\rP_\beta$.

The upper bound in Lemma \ref{Lem1} (i) gives the
so called ``slow rate'' of convergence for the Bregman divergence.
In Section 3, we provide ``fast rate'' of convergence for the Bregman divergence
via oracle inequalities for $|h_S|_1$ in (\ref{basic-ineq}).
The symmetrized Bregman divergence $\Delta(\hbeta,\beta^*)$
has the interpretations as the regret in prediction
error in linear regression, the symmetrized Kullback-Leibler (KL) divergence in
generalized linear models (GLM) and density estimation, and a spectrum loss
for the graphical Lasso, as shown in examples below.

\begin{example}
\label{Linear1}
\textbf{(Linear regression)}  Consider the linear regression model
\bel{linear-reg}
y_i = \sum_{j=1}^p x_{ij}\beta_j + \veps_i, \ \ i=1, \ldots, n,
\eel
where $y_i$ is the response variable, $x_{ij}$ are
predictors or design variables, and $\veps_i$  is the error term.
Let $y=(y_1, \ldots, y_n)'$ and let $X$ be the design matrix whose
$i$th row is $x^i=(x_{i1}, \ldots, x_{ip})$.
The estimator (\ref{Lasso}) is a weighted
Lasso with $\psi(\beta)=|X\beta|_2^2/(2n)$ and $z=X'y/n$ in (\ref{model}).
For predicting a vector $\ytil$ with $\rE_{\beta^*}[\ytil|X,y]=X\beta^*$,
\bes
n \Delta(\hbeta,\beta^*)
&=& |X\hbeta-X\beta^*|_2^2
\cr &=& \rE_{\beta^*}[|\ytil - X\hbeta|_2^2|X,y]
- \min_{\delta(X,y)}\rE_{\beta^*}[|\ytil-\delta(X,y)|_2^2|X,y]\}
\ees
is the regret of using the linear predictor $X\hbeta$ compared with the optimal predictor.
See \cite{GreenshteinR04} for several implications of (\ref{pred}).
\end{example} 

\begin{example}\label{logistic}
\textbf{(Logistic regression)}  We observe $(X,y)\in \real^{n\times(p+1)}$
with independent rows $(x^i,y_i)$, where $y_i\in\{0,1\}$ are binary response variables
with
\bel{logit-1}
\rP_\beta (y_i=1|x^i) = \pi_i(\beta) = \exp(x^i\beta)/(1+\exp(x^i\beta)) ,\ 1 \le i \le n.
\eel
The loss function (\ref{model}) is the average negative log-likelihood
\bel{logit-2}
\ell(\beta) = \psi(\beta)-z'\beta\ \hbox{ with }\
\psi(\beta) = \sum_{i=1}^n\frac{\log(1+\exp(x^i\beta))}{n},\ z=X'y/n.
\eel
Thus, (\ref{Lasso}) is a weighted $\ell_1$ penalized MLE.
For probabilities $\{\pi',\pi''\}\subset (0,1)$, the KL information is
$K(\pi',\pi'') = \pi'\log(\pi'/\pi'')+(1-\pi')\log\{(1-\pi')/(1-\pi'')\}$.
Since $\dpsi(\beta)= \sum_{i=1}^n x^i\pi_i(\beta)/n$ and
$\text{logit}(\pi_i(\beta^*))-\text{logit}(\pi_i(\beta))
=x^i(\beta^*-\beta)$, (\ref{Delta}) gives
\bes
\Delta(\beta,\beta^*) = \frac{1}{n}\sum_{i=1}^n
\Big\{K(\pi_i(\beta^*),\pi_i(\beta)) + K(\pi_i(\beta),\pi_i(\beta^*))\Big\}.
\ees
Thus, $\Delta(\beta^*,\beta)$ is the symmetrised KL-divergence.
\end{example} 

\begin{example}
\label{GLM1}
\textbf{(GLM).}
The GLM contains the linear and logistic regression models as special cases.
We observe $(X,y)\in \real^{n\times(p+1)}$ with rows $(x^i,y_i)$.
Suppose that conditionally on $X$, $y_i$ are independent under $\rP_\beta$ with
\bel{GLM}
y_i \sim f(y_i|\theta_i)
= \exp\Big(\frac{\theta_i y_i - \psi_0(\theta_i)}{\sigma^2} + \frac{c(y_i,\sigma)}{\sigma^2} \Big),\ \theta_i=x^i\beta.
\eel
Let $f_{(n)}(y|X,\beta)=\prod_{i=1}^nf(y_i| x^i\beta)$. The loss function
can be written as a normalized negative likelihood
$\ell(\beta)=(\sigma^2/n)\log f_{(n)}(y|X,\beta)$ with  $z=X'y/n$ and
$\psi(\beta) = \sum_{i=1}^n\{\psi_0(x^i\beta) + c(y_i,\sigma)\}/n$.
The KL divergence is
\bes
D\big(f_n(\cdot|X, \beta^*)\big\|f_n(\cdot|X, \beta)\big)
= \rE_{\beta^*} \log\Big(\frac{f_{(n)}(y|X,\beta^*)}{f_{(n)}(y|X,\beta)}\Big).
\ees
The symmetrized Bregman divergence can be written as
\bel{GLM-KL}
\Delta(\hbeta,\beta^*)
= \frac{\sigma^2}{n}\Big\{D\big(f_{(n)}(\cdot|X,\beta^*)\big\|f_{(n)}(\cdot|X,\hbeta)\big)
+D\big(f_{(n)}(\cdot|X,\hbeta)\big\|f_{(n)}(\cdot|X,\beta^*)\big)\Big\}.
\eel
\end{example}

\begin{example}
\label{nde1}
\textbf{(Nonparametric density estimation)} Although the focus of this paper is on regression models, here we illustrate that $\Delta(\hbeta, \beta^*)$ is the symmetrised KL divergence in the context of nonparametric density estimation. Suppose the
observations $y=(y_1,\ldots,y_n)'$ are iid from
$f(\cdot |\beta) = \exp\{ \langle \beta, T(\cdot)\rangle - \psi(\beta)\}$
under $\rP_\beta$, where $T(\cdot) = ( u_j(\cdot), j \le p)'$ with
certain basis functions $u_j(\cdot)$. Let the loss function $\ell(\beta)$ in (\ref{model}) be
the average negative log-likelihood $n^{-1}\sum_{i=1}^n \log f(y_i|\beta)$ with
$z = n^{-1}\sum_{i=1}^n T(y_i)$. Since $\rE_\beta T(y_i)=\dpsi(\beta)$,
the KL divergence is
\bes
D\big(f(\cdot|\beta^*)\big\|f(\cdot|\beta)\big)
= \rE_{\beta^*} \log\Big(\frac{f(y_i|\beta^*)}{f(y_i|\beta)}\Big)
= \psi(\beta)-\psi(\beta^*)-\langle \beta-\beta^*,\dpsi(\beta^*)\rangle.
\ees
Again, the
symmetrised KL divergence between the target density $f(\cdot|\beta^*)$
and the estimated density $f(\cdot|\hbeta)$ is
\bel{KL-div}
\Delta(\beta,\beta^*)=D\big(f(\cdot|\beta^*)\big\|f(\cdot|\hbeta)\big)
+D\big(f(\cdot|\hbeta)\big\|f(\cdot|\beta^*)\big).
\eel
\cite{vandeGeer08} pointed out that for this example, the natural choices of the
basis functions $u_j$ and weights $w_j$ satisfy $\int u_jd\nu=0$ and
$w_k^2 = \int u_k^2d\nu$.
\end{example}

\begin{example}
\label{glasso}
\textbf{(Graphical Lasso)} Suppose we observe $X\in \real^{n\times p}$
and would like to estimate the precision matrix
$\beta = (EX'X/n)^{-1}\in\real^{p\times p}$.
In the graphical Lasso, (\ref{model}) is the length normalized negative
likelihood with $\psi(\beta) = - \log\det\beta$, $z= - X'X/n$, and
$\langle \beta,z\rangle = - \trace(\beta z)$.
Since $\dpsi(\beta)=\rE_\beta z = - \beta^{-1}$, we find
\bel{trace-loss}
\Delta(\beta,\beta^*)=\trace\big((\hbeta-\beta^*)((\beta^*)^{-1}-\hbeta^{-1}\big)
=\sum_{j=1}^p(\lam_j-1)^2/\lam_j,
\eel
where $(\lam_1,\ldots,\lam_p)$ are the eigenvalues of
$(\beta^*)^{-1/2}\hbeta(\beta^*)^{-1/2}$. In graphical Lasso, the diagonal elements
are typically not penalized. Consider $\hw_{jk}=I\{j\neq k\}$, so that the penalty for
the off-diagonal elements are uniformly weighted. Since Lemma \ref{Lem1} requires
$|(z-\dpsi(\beta^*))_{jk}|\le\hw_{jk}\lam$, $\beta^*$ is taken to match $X'X/n$ on the
diagonal and the true $\beta^o$ in correlations.
Let $S=\{(j,k): \beta^o_{jk}\neq 0, j\neq k\}$. In the event
$\max_{j\neq k}|z_{jk}-\beta^*_{jk}|\le\lam$, Lemma \ref{Lem1} (i) gives
$\|(\beta^*)^{-1/2}\hbeta(\beta^*)^{-1/2} - I_{p\times p}\|_2=o(1)$ under
the condition $|S|\lam \max_{j\neq k}|\beta^*_{jk}|=o(1)$, where $\|\cdot\|_2$ is
the spectrum norm. \cite{RothmanBLZ08} proved the consistency of the graphical Lasso under
similar conditions with a different analysis.
\end{example}

\section{Oracle inequalities}
\label{oracle-ineq}
In this section, we extract upper bounds for the estimation error $\hbeta-\beta^*$
from the basic inequality (\ref{basic-ineq}). Since (\ref{basic-ineq}) is monotone
in the weights, the oracle inequalities are sharper when the weights $\hw_j$ are
smaller  in $S=\{j:\beta^*_j\neq 0\}$ and larger in $S^c$.

We say that a function $\phi(b)$ defined in
$\real^p$ is quasi star-shaped if $\phi(tb)$ is continuous and non-decreasing in $t\in [0,\infty)$
for all $b\in\real^p$ and $\lim_{b\to 0}\phi(b)=0$. All seminorms are quasi star-shaped.
The sublevel sets $\{b: \phi(b)\le t\}$ of a quasi star-shaped function are all star-shaped.
For $0\le \eta^*\le 1$ and any pair of quasi star-shaped functions
$\phi_0(b)$ and $\phi(b)$, define
\bel{GIF}
F(\xi,S; \phi_0,\phi) = \inf\Big\{
\frac{\Delta(\beta^*+b,\beta^*) e^{\phi_0(b)}}{|b_S|_1\phi(b)}:
b\in\scrC(\xi,S),\phi_0(b)\le\eta^*\Big\},
\eel
where $\Delta(\beta,\beta^*)$ is as in (\ref{Delta}). We refer to
$F(\xi,S; \phi_0,\phi)$ as a general invertibility factor (GIF) over the cone (\ref{cone}). 
The GIF plays a crucial role in developing
the error bounds for $\hbeta-\beta^*$. 
It extends the squared compatibility constant \cite{vandeGeerB09}
and the weak and sign-restricted cone invertibility factors
\cite{YeZ10} from the linear regression
model with $\phi_0(\cdot)=0$ to more general model (\ref{model}) and
from $\ell_q$ norms to general $\phi(\cdot)$.
They are all closely related to the restricted eigenvalues
\cite{BickelRT09, Koltchinskii09} as we will discuss in
Subsection 3.1.

The basic inequality  (\ref{basic-ineq}) implies that the symmetrized
Bregman divergence $\Delta(\hbeta,\beta^*)$ is no greater than
a linear function of $|h_S|_1$, where $h=\hbeta-\beta^*$.
If $\Delta(\hbeta,\beta^*)$ is no smaller than a linear function of the product
$|h_S|_1\phi(h)$, then an upper bound for $\phi(h)$ exists. 
Since the symmetrized Bregman divergence (\ref{Delta}) is approximately quadratic,
$\Delta(\hbeta,\beta^*)\approx h'\ddpsi(\beta^*)h$, in a neighborhood of $\beta^*$,
this is reasonable when $h=\hbeta-\beta^*$ is not too large and $\ddpsi(\beta^*)$
is invertible in the cone. A suitable factor $e^{\phi_0(b)}$ in (\ref{GIF})
forces the computation of this lower bound in a proper neighborhood of $\beta^*$.

We first provide a set of general oracle inequalities.

\begin{theorem}\label{th-est}
Let $\{z^*_0,z^*_1\}$ be as in (\ref{z^*}) with $S\supseteq\{j:\beta_j^*\neq 0\}$,
$\Omega_0$ in (\ref{Omega_0}),
$0\le \eta\le\eta^*\le 1$, and
$\{\phi_0(b),\phi(b)\}$ be a pair of quasi star-shaped functions.
Let $\phi_{1,S}(b)=|b_S|_1/|S|$. In the event
\bel{noise-cond}
\Omega_1 = \Omega_0\cap \Big\{\frac{|w_S|_\infty\lam+z^*_0}{\lam-z^*_1} \le \xi,\
\frac{|w_S|_\infty\lam+z^*_0}{F(\xi,S;\phi_0,\phi_0)}
\le \eta e^{-\eta}\Big\},
\eel
the following oracle inequalities hold:
\bel{oracle-1a}
&& \phi_0(\hbeta-\beta^*)\le \eta,\quad \phi(\hbeta-\beta^*)
\le \frac{e^\eta (|w_S|_\infty\lam+z^*_0)}{F(\xi,S;\phi_0,\phi)},
\\ \label{oracle-1b}
&& \Delta(\hbeta,\beta^*) + (\lam-z^*_1)|W_{S^c}(\hbeta-\beta^*)_{S^c}|_1
\le \frac{e^\eta (|w_S|_\infty\lam+z^*_0)^2|S|}{F(\xi,S;\phi_0,\phi_{1,S})}.
\eel
\end{theorem}

\begin{remark}
Sufficient conditions are given in Subsection 3.2 for (\ref{noise-cond})
to hold with high probability. See Lemma \ref{Lem2},
Remarks \ref{Remark3p1a} and \ref{Remark3p1} and Examples
\ref{example-linearB}, \ref{example-logisticB}, and \ref{example-loglinearB}.
\end{remark}

The oracle inequalities in Theorem \ref{th-est} control both the estimation
error in terms of $\phi_0(\hbeta-\beta^*)$ and the prediction error in terms
of the symmetrized Bregman divergence $\Delta(\hbeta,\beta^*)$ discussed
in Section 2.
Since they 
are based on (\ref{GIF})
in the intersection of the cone and the unit ball $\{b:\phi_0(b)\le 1/e\}$,
they are different from typical results in a small-ball analysis based on the
Taylor expansion of $\psi(\beta)$ at $\beta=\beta^*$.
Theorem~\ref{th-est} does allow
$\phi_0(\cdot)=0$ with $F(\xi,S;\phi_0,\phi_0)=\infty$ and $\eta=0$
in linear regression.

\subsection{The Hessian and related quantities}
We describe the relationship between the GIF (\ref{GIF}) and the Hessian of the
convex function $\psi(\cdot)$ in (\ref{model}) and examine cases where the
quasi star-shaped functions $\phi_0(\cdot)$ and $\phi(\cdot)$ are familiar seminorms.
Throughout, we assume that $\psi(\beta)$ is twice differentiable.
Let $\ddpsi(\beta)$ be the Hessian of $\psi(\beta)$ and $\Sigma^*=\ddpsi(\beta^*)$.

The GIF (\ref{GIF}) can be simplified if for a certain nonnegative-definite matrix $\Sigma$,
\bel{Hessian-cond-1}
\Delta(\beta^*+b,\beta^*) e^{\phi_0(b)} \ge \langle b, \Sigma b\rangle,\
\forall\ b\in\scrC(\xi,S),\ \phi_0(b) \le \eta^*.
\eel
Since $\Delta(\beta^*+h,\beta^*)=\int_0^1\langle h, \ddpsi(\beta^*+th)h\rangle dt$
by (\ref{Delta}), (\ref{Hessian-cond-1}) is a smoothness condition on the Hessian
when $\Sigma=\Sigma^*$. In what follows, $\Sigma=\Sigma^*$ is allowed in all
statements unless otherwise stated.
Under (\ref{Hessian-cond-1}), (\ref{GIF}) is bounded from below by the simple GIF,
\bel{GIF-1}
F_0(\xi,S;\phi) = \inf_{b\in\scrC(\xi,S)}
\frac{\langle b, \Sigma b\rangle}{|b_S|_1\phi(b)}.
\eel
In linear regression, $F_0(\xi,S;\phi)$ is the square of
the compatibility factor for $\phi(b)=\phi_{1,S}(b)=|b_S|_1/|S|$
\cite{vandeGeer07} and the cone invertibility factor for
$\phi(b)=\phi_q(b)=|b|_q/|S|^{1/q}$ \cite{YeZ10}.
They are both closely related to the restricted isometry property (RIP) \cite{CandesT05},
the sparse Rieze condition (SRC) \cite{ZhangH08}, and the restricted eigenvalue
\cite{BickelRT09}. Extensive discussion of these quantities can be found in
\cite{BickelRT09, vandeGeerB09, YeZ10}.
The following corollary is an extension of an oracle inequality of
\cite{YeZ10} for the linear regression model.

\begin{corollary}\label{Cor1} Let 
$\eta\le\eta^*\le 1$.
Suppose (\ref{Hessian-cond-1}) holds. Then, in the event
\bes
\Omega_0\cap\big\{ |w_S|_\infty\lam+z^*_0\le\min\big(\xi(\lam-z^*_1),
\eta e^{-\eta}F_0(\xi,S;\phi_0)\big)\big\},
\ees
(\ref{oracle-1a}) and (\ref{oracle-1b}) hold with
$F(\xi,S;\phi_0,\phi)$ replaced by the simpler
$F_0(\xi,S;\phi)$ in (\ref{GIF-1}). In particular, in the same event,
\bel{oracle-2}
\phi_0(h)\le\eta,\
|h|_q\le \frac{e^\eta (|w_S|_\infty\lam+z^*_0)|S|^{1/q}}{F_0(\xi,S;\phi_q)},\ \forall\, q > 0,
\eel
with $\phi_q(b)=|b|_q/|S|^{1/q}$ and $h=\hbeta-\beta^*$, and
with $\phi_{1,S}(b)=|b_S|_1/|S|$,
\bel{oracle-3}
e^{-\eta} h'\Sigma h
\le \Delta(\hbeta,\beta^*)
\le \frac{e^{\eta} (|w_S|_\infty\lam+z^*_0)^2|S|}{F_0(\xi,S;\phi_{1,S})} -
(\lam-z^*_1)|W_{S^c}h_{S^c}|_1.
\eel
\end{corollary}

Here the only differences between the general model (\ref{model}) and linear
regression ($\phi_0(b)=0$) are the extra factor $e^\eta$ with $\eta\le 1$, the extra constraint
$|w_S|_\infty\lam+z^*_0\le \eta e^{-\eta}F_0(\xi,S;\phi_0)$, and the extra
condition (\ref{Hessian-cond-1}).
Moreover, (\ref{GIF-1}) explicitly expresses all conditions
on $F_0(\xi,S;\phi)$ as properties of a fixed $\Sigma$.

\medskip
\begin{example}
\label{linear-cont1}
\textbf{(Linear regression: oracle inequalities).}
For $\psi(\beta)=|Xb|_2^2/(2n)$ and $\Sigma=X'X/n$,
$F_0(\xi,S;\phi_q)$ is the weak cone invertibility factor \cite{YeZ10} and
$F^{1/2}_0(\xi,S;\phi_{1,S})$ is the compatibility constant \cite{vandeGeer07}
\bel{kappa_*}
\kappa_*(\xi,S) = \inf_{b\in \scrC(\xi,S)} \frac{|S|^{1/2}|X b|_2}{|b_S|_1n^{1/2}}
= \inf_{b\in \scrC(\xi,S)} \Big(\frac{b'\Sigma b}{|b_S|_1^2/|S|} \Big)^{1/2}.
\eel
They are all closely related to
the $\ell_2$ restricted eigenvalues
\bel{RE_2}
RE_2(\xi,S) = \inf_{b\in \scrC(\xi,S)} \frac{|X b|_2}{|b|_2n^{1/2}}
= \inf_{b\in \scrC(\xi,S)} \Big(\frac{b'\Sigma b}{|b|_2^2} \Big)^{1/2}
\eel
\cite{BickelRT09, Koltchinskii09}.
Since $|b_S|_1^2\le |b|_2^2|S|$, $\kappa_*(\xi,S)\ge RE_2(\xi,S)$
\cite{vandeGeerB09}.
For the Lasso with $\hw_j=1$,
\bel{SCIF}
|\hbeta-\beta^*|_2
\le \frac{|S|^{1/2}(\lam+z^*_0)}{SCIF_2(\xi,S)}
\le \frac{|S|^{1/2}(\lam+z^*_0)}{F_0(\xi,S;\phi_2)}
\le \frac{|S|^{1/2}(\lam+z^*_0)}{\kappa_*(\xi,S)RE_2(\xi,S)}
\eel
in the event $\lam+z^*_0\le \xi(\lam-z^*_1)$ \cite{YeZ10}, where
\bes
SCIF_q(\xi,S) = \inf_{b\in \scrC_-(\xi,S)}|\Sigma b|_\infty/\phi_q(b),\
\phi_q=|b|_q/|S|^{1/q}.
\ees
Thus, cone and general invertibility factors yield sharper $\ell_2$ oracle inequalities.
\end{example}

The factors in the oracle inequalities in (\ref{SCIF}) do not have the same
order for large $|S|$ and certain design matrices $X$.
Although the oracle inequality based on $SCIF_2(\xi,S)$ is the sharpest in
(\ref{SCIF}), it seems not to lead to a simple extension to the general convex
minimization with (\ref{model}). Thus, we settle with extensions of the
second sharpest oracle inequality in (\ref{SCIF}) with $F_0(\xi,S;\cdot)$.

\subsection{Oracle inequalities for the Lasso in GLM}

An important special case of the general formulation is the $\ell_1$-penalized
estimator in a generalized linear model (GLM) \cite{McCullaghN89}.
This is Example \ref{GLM1} in Subsection 2.2, where we set up the notation
in (\ref{GLM}) and gave the KL divergence interpretation to (\ref{Delta}).
The $\ell_1$ penalized, normalized negative likelihood is
\bel{GLMfz}
\ell(\beta)=\psi(\beta)-z'\beta,\ \hbox{ with }
\psi(\beta) = C_n(y,\sigma)+\sum_{i=1}^n \frac{\psi_0(x^i\beta)}{n} \hbox{ and } z = \frac{X'y}{n}.
\eel
Assume that $\psi_0$ is twice differentiable. Denote the first and second
derivatives of $\psi_0$ by $\dpsi_0$ and $\ddpsi_0$, respectively.
The gradient and Hessian are
\bel{GLM-1}
\dpsi(\beta) = X'\dpsi_0(\theta)/n \  \mbox{ and }\
\ddpsi(\beta) = X'\diag(\ddpsi_0(\theta))X/n,
\eel
where $\theta=X\beta$ and $\dpsi_0$ and $\ddpsi_0$
are applied to the individual components of $\theta$.

A crucial condition in our analysis of the Lasso in GLM is
\bel{cond-6}
\max_{i\le n}\Big|\log\Big(\frac{\ddpsi_0(x^i\beta^* + t)}{\ddpsi_0(x^i\beta^*)}\Big)\Big|
\le {M}_1|t|, \ \forall M_1|t|\le{\eta^*}
\eel
where ${M}_1$ and ${\eta^*}$ are constants determined by $\psi_0$.
This condition gives
\bes
\Delta(\beta^*+b,\beta^*)  = \int_0^1\langle b,\ddpsi(\beta^*+tb)b\rangle dt
\ge \int_0^1 \sum_{tM_1|x^ib|\le\eta^*}\frac{\ddpsi_0(x^i\beta^*)(x^ib)^2}{n e^{tM_1|x^ib|}}dt,
\ees
which implies the following lower bound for the GIF in (\ref{GIF}):
\bel{GIF-bound}
F(\xi,S;\phi_0,\phi) 
&\ge& \inf_{b\in\scrC(\xi,S),\phi_0(b)\le\eta^*}\sum_{i=1}^n
\frac{\ddpsi_0(x^i\beta^*)(x^ib)^2}{n|b_S|_1\phi(b)}\int_0^1 I\{tM_1|x^ib|\le \phi_0(b)\}dt
\cr &=& \inf_{b\in\scrC(\xi,S),\phi_0(b)\le\eta^*}\sum_{i=1}^n
\frac{\ddpsi_0(x^i\beta^*)}{n|b_S|_1\phi(b)}\min\Big(\frac{|x^ib|\phi_0(b)}{M_1},(x^ib)^2\Big), 
\eel
due to $(x^ib)^2\int_0^1 I\{tM_1|x^ib|\le\phi_0(b)\}dt
= \min\{|x^ib|\phi_0(b)/M_1,(x^ib)^2\}$. For seminorms $\phi_0$ and $\phi$,
the infimum above can be taken over a fixed value of $\phi_0(b)$ due to scale
invariance. Thus, for $\phi_0(b)=M_2|b|_2$ and seminorms $\phi$,
the lower bound in (\ref{GIF-bound}) is
\bel{GIF-2}
F^*(\xi,S;\phi) = \inf_{b\in\scrC(\xi,S),|b|_2=1}\sum_{i=1}^n
\frac{\ddpsi_0(x^i\beta^*)}{n|b_S|_1\phi(b)}
\min\Big(\frac{|x^ib|M_2}{M_1},(x^ib)^2\Big). 
\eel
If (\ref{cond-6}) holds with $\eta^*=\infty$, the convexity of $e^{-t}$
yields (\ref{Hessian-cond-1}) with
\bel{phi_0-1}
\phi_0(b) = \frac{M_1\sum_{i=1}^n\ddpsi_0(x^i\beta^*) |x^ib|^3}
{\sum_{i=1}^n \ddpsi_0(x^i\beta^*)(x^ib)^2}\le M_1|Xb|_\infty,
\eel
with an application of the Jensen inequality.
This gives a special $F_0(\xi,S;\phi_0)$ as
\bel{GIF-3}
F_*(\xi,S) = \inf_{b\in\scrC(\xi,S)}\frac{n \langle b, \Sigma^* b\rangle^2/(M_1|b_S|_1)}
{\sum_{i=1}^n\ddpsi_0(x^i\beta^*)|x^ib|^3}.
\eel
We note that since
$|Xb|_\infty\le |X_S|_\infty|b_S|_1+|X_{S^c}W_{S^c}^{-1}|_\infty|W_{S^c}b_{S^c}|
\le \{|X_S|_\infty+\xi |X_{S^c}W_{S^c}^{-1}|_\infty\} |b_S|$ in the cone $\scrC(\xi,S)$ in
(\ref{cone}), for $\phi_0(b) = M_3|b_S|_1$ with
$M_3=M_1\{|X_S|_\infty+\xi |X_{S^c}W_{S^c}^{-1}|_\infty\}$,
(\ref{Hessian-cond-1}) automatically implies the stronger
\bel{phi_0-2}
e^{-\phi_0(b)}\langle b, \Sigma^* b\rangle
\le \Delta(\beta^*+b,\beta^*) \le e^{\phi_0(b)}\langle b, \Sigma^* b\rangle,\
\forall\ b\in\scrC(\xi,S),\
\phi_0(b) \le{\eta^*}.
\eel

Under condition (\ref{cond-6}), we may also use the following large deviation inequalities
to find explicit penalty levels to guarantee (\ref{noise-cond}).

\begin{lemma}\label{Lem2}
(i) Suppose (\ref{GLM}) and (\ref{cond-6}) hold with certain $\{M_1,{\eta^*}\}$ and
the $w_j$ in (\ref{Omega_0}) are deterministic.
Let $x_j$ be the columns of $X$, $\Sigma^*_{ij}$
be the elements of $\Sigma^*=\ddpsi(\beta^*)$. For positive constants
$\{\lam_0,\lam_1\}$ define $t_j=\lam_0I\{j\in S\}+w_j\lam_1I\{j\not\in S\}$.
Suppose
\bel{lm-2-cond-1}
M_1 \max_{j\le p}\big(|x_j|_\infty |t_j/\Sigma^*_{jj}\big) \le \eta_0e^{\eta_0}\ \hbox{ and }\
\sum_{j=1}^p \exp\Big\{ - \frac{n t_j^2e^{-\eta_0}}{2\sigma^2\Sigma^*_{jj}}\Big\} \le \frac{\eps_0}{2}
\eel
for certain constants $\eta_0\le {\eta^*}$ and $\eps_0>0$.
Then, $\rP_{\beta^*}\Big\{ z^*_0\le\lam_0, z^*_1\le \lam_1\Big\} \ge 1- \eps_0$. \\
(ii) If $c_0=\max_t\ddpsi(t)$, then part (i) is still valid if (\ref{cond-6})
and (\ref{lm-2-cond-1}) are replaced by
\bel{lm-2-cond-2}
\sum_{j=1}^p \exp\Big\{ - \frac{n^2 t_j^2}{2\sigma^2c_0|x_j|_2^2}\Big\} \le \frac{\eps_0}{2}.
\eel
In particular, if $|x_j|_2^2=n,1\le j \le p,  w_j=1, j \not\in S $ and $\lam_0=\lam_1=\lam$ (so
$t_j=\lam$), then part (i) still holds if $\lam \ge \sigma \sqrt{(2 c_0/n) \log(2p/\eps_0)}$.
\end{lemma}

The following theorem is a consequence of Theorem \ref{th-est},
Corollary \ref{Cor1} and Lemma \ref{Lem2}.

\begin{theorem}\label{th-est-GLM}
(i) Let $\hbeta$ be the Lasso (\ref{Lasso}) with the loss function in (\ref{GLMfz}).
Let $\beta^*$ be a target vector and $h=\hbeta-\beta^*$.
Suppose (\ref{GLM}) and (\ref{cond-6}) hold with certain $\{M_1,\eta^*\}$.
Let $F^*(\xi,S;\phi)$ be as in (\ref{GIF-2}) with $S=\{j:\beta^*_j\neq 0\}$ and a constant $M_2$.
Let $\eta\le 1\wedge\eta^*$ and $\{\lam,\lam_0,\lam_1\}$ satisfy
\bel{th-est-GLM-1}
|w_S|_\infty\lam+\lam_0\le\min\big\{\xi(\lam-\lam_1),\eta e^{-\eta}F^*(\xi,S;M_2|\cdot|_2)\big\}.
\eel
Then, in the event $\Omega_0\cap\big\{\max_{k=0,1} \big(z^*_k/\lam_k\big) \le 1\big\}$
with the $z^*_k$ in (\ref{z^*}) and $\Omega_0$ in (\ref{Omega_0}),
\bel{th-est-GLM-2}
\Delta(\beta^*+ h,\beta^*)
\le \frac{e^{\eta}(|w_S|_\infty\lam+\lam_0)^2|S|}{F^*(\xi,S;\phi_{1,S})},\quad
\phi(h) \le \frac{e^\eta (|w_S|_\infty\lam+\lam_0)}{F^*(\xi,S;\phi)}
\eel
for all seminorms $\phi$.
Moreover, if either (\ref{lm-2-cond-1}) or (\ref{lm-2-cond-2}) holds for the $\{\lam_0,\lam_1\}$ and
$W$ is deterministic, then
\bes
\rP_{\beta^*}\big\{\hbox{(\ref{th-est-GLM-2}) holds for all seminorms }
\phi \big\}\ge \rP_{\beta^*}(\Omega_0)-\eps_0.
\ees
(ii) If $\eta^*=\infty$ and (\ref{th-est-GLM-1}) holds with $F^*(\xi,S;M_2|\cdot|_2)$
replaced by the $F_*(\xi,S)$ in (\ref{GIF-3}), then the conclusions of part (i) hold
with $F^*(\xi,S;\cdot)$ replaced by the $F_0(\xi,S;\cdot)$ in (\ref{GIF-1}). Moreover,
(\ref{th-est-GLM-2}) can be strengthened with the lower bound
$\Delta(\beta^*+ h,\beta^*)\ge e^{-\eta}\langle h,\Sigma^* h\rangle$.
\\ (iii) For any ${\eta^*}>0$, the conclusions of part (ii) hold if $F_*(\xi,S)$ is replaced by
$\kappa_*^2(\xi,S)/(M_3|S|)$ in (\ref{th-est-GLM-1})
with the $M_3$ in (\ref{phi_0-2}).
\end{theorem}

\begin{remark}
\label{Remark3p0}
Since $\phi=\phi_0$ is allowed in (\ref{th-est-GLM-2}),
(\ref{th-est-GLM-2}) implies $\phi_0(h)\le\eta$ with $\phi_0(h)=M_2|h|_2$ in part (i)
and the $\phi_0$ in (\ref{phi_0-1}) in part (ii).
Similarly, under the conditions of Theorem \ref{th-est-GLM} (iii),
$M_3|h_S|_1\le\eta\le {\eta^*}$, so that (\ref{phi_0-2}) holds
with $b=h=\hbeta-\beta^*$.
\end{remark}

\begin{remark}
\label{Remark3p1a}
If either (\ref{lm-2-cond-1}) or (\ref{lm-2-cond-2}) holds for $\{\lam_0,\lam_1\}$ and
$W$ is deterministic, then (\ref{th-est-GLM-2}) implies
$\rP_{\beta^*}\{$(\ref{noise-cond}) holds$\} \ge \rP_{\beta^*}(\Omega_0)-\eps_0$.
\end{remark}

\begin{remark}
\label{Remark3p1}
Suppose $\{\min_{j\not\in S}w_j, \min_j\Sigma^*_{jj}\}$
are bounded away from zero, \\ $\{\max_{j\in S}w_j, \max_j\Sigma^*_{jj}, M_1\}$
are bounded, and $\{1+F_*^2(\xi,S)\}(\log p)/n \to 0$.
Then, (\ref{lm-2-cond-1}) holds with $\lam_0=\lam_1=a \sigma \sqrt{(2/n)\log(p/\eps_0)}$
for certain $a\le (1+o(1))\max_j(\Sigma^*)_{jj}^{1/2}/w_j$,
due to $\max\{\lam_0, \eta, \eta_0\} \to 0+$. Again, the conditions and conclusions of
Theorem \ref{th-est-GLM} ``converge'' to those for the linear regression
as if the Gram matrix is $\Sigma^*$.
\end{remark}

\begin{remark}
\label{Remark3p2}
In Theorem \ref{th-est-GLM},
the key condition (\ref{th-est-GLM-1}) is weaker in parts (i) and (ii) than part (iii),
although part (ii) requires ${\eta^*}=\infty$.
For $\Sigma=\Sigma^*$ and $M_1=M_2 \le M_3/(1+\xi)$,
\bes
\kappa_*^2(\xi,S)/(M_3|S|) \le \min\Big\{ F_*(\xi,S), F^*(\xi,S;M_2|\cdot|_2)\Big\},
\ees
since $n^{-1}\sum_{i=1}^n \ddpsi_0(x^i\beta^*)|x^ib|^3/\langle b,\Sigma^* b\rangle
\le |Xb|_\infty \le |b_S|_1M_3/M_1$ as in the derivation of (\ref{phi_0-2})
and $|b|_2\le (1+\xi)|b_S|_1$ in the cone (\ref{cone}).
For the more familiar $\kappa_*^2(\xi,S)/(M_3|S|)$,
(\ref{th-est-GLM-1}) essentially requires a small $|S|\sqrt{(\log p)/n}$.
The sharper Theorem \ref{th-est-GLM} (i) and (ii) provides conditions to
relax the requirement to a small $|S|(\log p)/n$.
\end{remark}

\begin{remark}
\label{Remark3p3}
For $\hw_j=1$, \cite{NegahbanRWY10} considered 
M-estimators under a
restricted strong convexity condition. For the GLM, they considered iid
sub-Gaussian $x^i$ and used empirical process theory to bound
$\Delta(\beta^*+b,\beta^*)/\{|b|_2(|b|_2-c_0|b|_1\}$ from below
over the cone (\ref{cone}) with a small $c_0$. Their result extends the
$\ell_2$ error bound $|S|^{1/2}(\lam+z^*_0)/RE_2^2(\xi,S)$ of
\cite{BickelRT09},
while Theorem~\ref{th-est-GLM} extends
the sharper (\ref{SCIF}) with the factor $F_0(\xi,S;\phi_2)$.
Theorem \ref{th-est-GLM} applies to both deterministic and random designs.
Similar to \cite{NegahbanRWY10}, for iid sub-Gaussian $x^i$,
empirical process theory can be use to verify (\ref{th-est-GLM-1})
with $F^*(\xi,S;M_2|\cdot|_2)\gtrsim |S|^{-1/2}$,
provided that $|S|(\log p)/n$ is small.
\end{remark}

\begin{example}
\label{example-linearB}
\textbf{(Linear regression: oracle inequalities, continuation)}
For the linear regression model (\ref{linear-reg}) with quadratic loss,
$\psi_0(\theta)=\theta^2/2$, so that
(\ref{cond-6}) holds with $M_1=0$ and ${\eta^*}=\infty$.
It follows that $F^*(\xi,S;M_2|\cdot|_2)=\infty$
and (\ref{th-est-GLM-1}) has the interpretation with $\eta=0+$ and
$\eta e^{-\eta}F^*(\xi,S;M_2|\cdot|_2)=\infty$.
Moreover, since $M_1=0$, $\eta_0=0+$ in (\ref{lm-2-cond-1}).
Thus, the conditions and conclusions of Theorem \ref{th-est-GLM} ``converge''
to the case of linear regression as $M_1\to 0+$.
Suppose $\veps_i\sim N(0,\sigma^2)$ as in (\ref{GLM}).
 For $\hw_j=w_j=1$ and $\Sigma^*_{jj}=\sum_{i=1}^nx_{ij}^2/n=1$, (\ref{lm-2-cond-1}) holds
with $\lam_0=\lam_1=\sigma\sqrt{(2/n)\log(p/\eps_0)}$ and (\ref{th-est-GLM-1}) holds
with $\lam = \lam_0(1+\xi)/(1-\xi)$.
The value of $\sigma$ can be estimated iteratively
using the mean residual squares \cite{StadlerBG10, SunZ11}.
Alternatively, cross-validation can be used to pick $\lam$.
For $\phi(b)=\phi_2(b)=|b|_2/|S|^{1/2}$,
(\ref{th-est-GLM-2}) matches (\ref{SCIF}) with the factor $F_0(\xi,S;\phi_2)$.
\end{example}

\begin{example}
\label{example-logisticB}
\textbf{(Logistic regression: oracle inequalities)}
The model and loss function are given in (\ref{logit-1}) and (\ref{logit-2})
respectively. Here we verify the conditions of Theorem \ref{th-est-GLM}.
Condition (\ref{cond-6}) holds with $M_1=1$ and ${\eta^*}=\infty$;
Since $\psi_0(t)=\log(1+e^t)$,
\bes
\frac{\ddpsi_0(\theta+t)}{\ddpsi_0(\theta)}
= \frac{e^{t}(1+e^\theta)^2}{(1+e^{\theta+t})^2}
\ge \begin{cases}e^{-|t|} & t<0 \cr
e^{-t}(1+e^\theta)^2/(e^{-t}+e^\theta)^2\ge e^{-|t|} & t>0.
\end{cases}
\ees
Since $\max_t \ddpsi(t) = c_0=1/4$ we can apply (\ref{lm-2-cond-2}). In particular,
if $\hw_j=w_j=1=|x_j|_2^2/n$, $\lam = \{(\xi+1)/(\xi-1)\}\sqrt{(\log(p/\eps_0))/(2n)}$ and
$\lam\{2\xi/(\xi+1)\}/F_*(\xi,S)\le\eta e^{-\eta}$, then (\ref{th-est-GLM-2}) holds with at least probability $1-\eps_0$ under $\rP_{\beta^*}$.
For such $\What$ and $X$, an adaptive choice of the penalty level is
$\lam = \hsigma\sqrt{(2/n)\log p}$
with $\hsigma^2=\sum_{i=1}^n\pi_i(\hbeta)\{1-\pi_i(\hbeta)\}/n$,
where $\pi_i(\beta)$ is as in Example \ref{logistic}.
\end{example}

\begin{example}
\label{example-loglinearB}
\textbf{(Log-linear models: oracle inequalities)}
 Consider counting data with $y_i\in \{0, 1, 2,...\}$.
In log-linear models, it is assume that
\bel{log-linear-1}
\rE_\beta(y_i) = e^{\theta_i},\ \theta_i = x^i\beta, \ 1\le i \le n.
\eel
The average negative Poisson log-likelihood function is
\bel{log-linear-2}
\ell(\beta) = \psi(\beta)-z'\beta,\
\psi(\beta) = \sum_{i=1}^n \frac{\exp(x^i\beta)-\log(y_i!)}{n},\ z=X'y/n.
\eel
Again this is a GLM.
In this model, $\psi_0(t)=e^t$, so that (\ref{cond-6}) holds with
$M_1=1$ and ${\eta^*}=\infty$.
Although (\ref{lm-2-cond-2}) is not useful with $c_0=\infty$,
(\ref{lm-2-cond-1}) can be used in Theorem \ref{th-est-GLM}.
\end{example}

\section{Adaptive and multistage methods}
\label{multistage-section}

We consider in this section an adaptive Lasso and its repeated applications, with weights recursively generated based a concave penalty function. 
This approach appears to provide the most appealing choice of weights both from heuristic and theoretical standpoints.
The analysis here is based on the results in Section 3 and
the main idea in \cite{Zhang10-multistage}.

Let $\rho_\lam(t)$ be a penalty function with $\drho_\lam(0+)=\lam$, where
$\drho_\lam(t)=(\pa/\pa t)\rho_\lam(t)$. Define
\bel{kappa}
\kappa = \sup_{0<t_1<t_2}\frac{|\drho_\lam(t_2)-\drho_\lam(t_1)|}{t_2-t_1}.
\eel
Let $\Sigma$ be as in (\ref{Hessian-cond-1}) and $\scrC(\xi,S)$ be the cone in (\ref{cone}).
Define
\bel{F_2}
F_2(\xi,S) = \inf\Big\{\frac{b'\Sigma b}{|b_S|_2|b|_2}: 0\neq b\in\scrC(\xi,S)\Big\}.
\eel
The quantity $F_2(\xi,S)$ is slightly larger than the square of the restricted eigenvalue
(\ref{RE_2}) for a design matrix $X$ when $\Sigma = X'X/n$.
Given $0<\eps_0<1$,
the components of the error vector $z-\dpsi(\beta^*)$ are sub-Gaussian if
for all $0\le t\le \sigma\sqrt{(2/n)\log(4p/\eps_0)}$,
\bel{subGaussian}
\rP_{\beta^*}\Big\{ |(z-\dpsi(\beta^*))_j| \ge t\Big\}
\le 2 e^{ - nt^2/(2\sigma^2)}.
\eel
This condition holds for all GLM when the components of $X\beta^*$ are uniformly
in the interior of the natural parameter space for the exponential family.

\begin{theorem}\label{th-adaptive-Lasso}
Suppose (\ref{Hessian-cond-1}) holds.
Let $\kappa$ be as in (\ref{kappa}),
$S_0=\{j:\beta_j^*\neq 0\}$, $\lam_0>0$, $0<\eta<1$, $0<\gamma_0<1/\kappa$, $A>1$,
and $\xi\ge (A+1)/(A-1)$. 
Suppose
\bel{th-adaptive-Lasso-1}
\lam_0\{1+A/(1-\kappa\gamma_0)\} \le F_0(\xi,S;\phi_0)\eta e^{-\eta},\
F_*\le F_2(\xi,S),
\eel
for all $S\supseteq S_0$ with $|S\setminus S_0|\le \ell^*$,
where $F_0(\xi,S;\phi_0)$ is as in (\ref{GIF-1}) and $F_2(\xi,S)$ as in (\ref{F_2}).
Let $\tbeta$ be an initial estimator of $\beta$ and $\hbeta$ be as in (\ref{Lasso}) with
$\what_j = \drho_\lam(|\tbeta_j|)/\lam$ and $\lam = A\lam_0/(1-\kappa\gamma_0)$.
Then,
\bes
|\hbeta-\beta^*|_2
\le \frac{e^\eta}{F_*}\Big\{|\drho_\lam(|\beta^*_{S_0}|)|_2 + |\{z-\dpsi(\beta^*)\}_{S_0}|_2
 + \Big(\kappa+\frac{1}{\gamma_0A}-\frac{\kappa}{A}\Big) |\tbeta-\beta^*|_2\Big\}
\ees
in the event $\{|(\tbeta-\beta)_{S_0^c}|_2^2\le\gamma_0^2\lam^2\ell^*\}
\cap\{|z-\dpsi(\beta^*)|_\infty\le\lam_0\}$.
Moreover, if (\ref{subGaussian}) holds and
$\lam_0=\sigma\sqrt{(2/n)\log(2p/\eps_0)}$ with $0<\eps_0<1$, then
$\rP_{\beta^*}\big\{ |z-\dpsi(\beta^*)| \ge \lam_0\big\}\le\eps_0$.
\end{theorem}

Theorem \ref{th-adaptive-Lasso} raises the possibility that $\hbeta$
improves $\tbeta$ under proper conditions.  Thus it is desirable to repeatedly apply this adaptive Lasso in the following way,
\bel{recursive-Lasso}
\hbeta^{(k+1)} = \argmin_{\beta}\Big\{\ell(\beta)
+\sum_{j=1}^p \drho_\lam(\hbeta_j^{(k)})|\beta_j|\Big\}, k=0, 1, \ldots . 
\eel
Such multistage algorithms have been considered in
\cite{FanL01,ZouL08,Zhang10-multistage}. As discussed in Remark \ref{th-recursive-Lasso-remark} below,
it is beneficial to use a concave penalty $\rho_{\lam}$ 
in (\ref{recursive-Lasso}). Natural choices of $\rho_{\lam}$ include the
smoothly clipped absolute deviation and minimax concave penalties
\cite{FanL01, Zhang10-mc+}. 

\begin{theorem}\label{th-recursive-Lasso}
Let $\{\kappa,S_0,\lam_0,\eta,\gamma_0,A,\xi,\ell^*,\lam\}$ be the same as
Theorem \ref{th-adaptive-Lasso}.
Let $\hbeta^{(0)}$ be the unweighted Lasso with $\what_j=1$ in (\ref{Lasso})
and $\hbeta^{(\ell)}$ be the $\ell$-th iteration of the recursion (\ref{recursive-Lasso})
initialized with $\hbeta^{(0)}$. Let $F_0(\xi,S_0;\phi_2)$ be the simple GIF in (\ref{GIF-1})
with $\phi_2(h)=|h|_2/|S|^{1/2}$.
Suppose (\ref{th-adaptive-Lasso-1}) holds and
\bel{th-recursive-Lasso-cond-1}
e^\eta\{1+(1-\kappa\gamma_0)/A\}\sqrt{|S_0|}/F_0(\xi,S_0;\phi_2)
\le \gamma_0\sqrt{\ell^*}.
\eel
Define $r_0 = (e^\eta/F_*)\{\kappa+1/(\gamma_0A)-\kappa/A\}$.
Suppose $r_0<1$. Then,
\bel{th-recursive-Lasso-1}
|\hbeta^{(\ell)}-\beta^*|_2
\le \frac{|\drho_\lam(|\beta^*_{S_0}|)|_2 + |\{z-\dpsi(\beta^*)\}_{S_0}|_2}
{e^{-\eta}F_*(1-r_0)/(1-r_0^\ell)}
 + \frac{r_0^\ell e^\eta\lam\{1+(1-\kappa\gamma_0)/A\}}{F_0(\xi,S_0;\phi_2)/|S_0|^{1/2}}
\eel
in the event
\bel{th-recursive-Lasso-event}
\Big\{|z-\dpsi(\beta^*)|_\infty\le\lam_0\Big\}\cap
\Big\{\frac{|\drho_\lam(|\beta^*_{S_0}|)|_2 + |\{z-\dpsi(\beta^*)\}_{S_0}|_2}
{e^{-\eta}F_*(1-r_0)} \le \gamma_0\lam\sqrt{\ell^*}\Big\}.
\eel
Moreover, if (\ref{subGaussian}) holds and
$\lam_0=\sigma\sqrt{(2/n)\log(4p/\eps_0)}$ with $0<\eps_0<1$,
then the intersection of the events (\ref{th-recursive-Lasso-event}) and
$\{|\{z-\dpsi(\beta^*)\}_{S_0}|_2\le n^{-1/2}\sigma\sqrt{2|S_0|\log(4|S_0|/\eps_0)}\}$
happens with at least $\rP_{\beta^*}$ probability $1-\eps_0$, provided that
\bel{th-recursive-Lasso-cond-2}
\frac{|\drho_\lam(|\beta^*_{S_0}|)|_2 + n^{-1/2}\sigma\sqrt{2|S_0|\log(4|S_0|/\eps_0)\}}}
{e^{-\eta}F_*(1-r_0)} \le \frac{\gamma_0A \lam_0\sqrt{\ell^*}}{1-\kappa\gamma_0}.
\eel
\end{theorem}

\begin{remark}\label{th-recursive-Lasso-remark}
Define
$R^{(0)} = e^\eta\lam\{1+(1-\kappa\gamma_0)/A\}|S_0|^{1/2}/F_0(\xi,S_0;\phi_2)$ and
\bes
R^{(\infty)} = \frac{|\drho_\lam(|\beta^*_{S_0}|)|_2 + |\{z-\dpsi(\beta^*)\}_{S_0}|_2}
{e^{-\eta}F_*(1-r_0)},\ R^{(\ell)} = (1-r_0^\ell)R^{(\infty)}+ r_0^\ell R^{(0)},
\ees
as in the right-hand side of (\ref{th-recursive-Lasso-1}).
Theorem \ref{th-recursive-Lasso} asserts that
$|\hbeta^{(\ell)}-\beta^*|\le 2R^{(\infty)}$ after
$\ell = |\log r_0|^{-1}\log(R^{(\infty)}/R^{(0)})$ iterations of the recursion
(\ref{recursive-Lasso}). Under condition (\ref{subGaussian}),
\bes
\rE_{\beta^*}R^{(\infty)} \le \{|\drho_\lam(|\beta^*_{S_0}|)|_2 + 2\sigma\sqrt{|S_0|/n}\}
e^{\eta}/\{F_*(1-r_0)\}.
\ees
Suppose $\rho_\lam(t)$ is concave in $t$, then
 $|\drho_\lam(|\beta^*_{S_0}|)|_2\le \drho_\lam(0+)|S_0|^{1/2}=\lam|S_0|^{1/2}$.
This component of $\rE_{\beta^*}R^{(\infty)}$
matches the noise inflation due to model selection since
$\lam\asymp \lam_0 =\sigma \sqrt{(2/n)\log(p/\eps_0)}$.
This noise inflation diminishes when $\min_{j\in S_0}|\beta_j^*|\ge \gamma \lam$
when $\drho_\lam(t)=0$ for $|t|\ge \gamma\lam$, yielding the super-efficient error bound
$\rE_{\beta^*}R^{(\infty)} \le \{2\sigma\sqrt{|S_0|/n}\}e^{\eta}/\{F_*(1-r_0)\}$. This risk bound
$R^{(\infty)}$ is comparable with those for concave penalized least squares
in linear regression \cite{Zhang10-mc+}.
\end{remark}

\begin{remark} For $\log(p/n)\asymp \log p$, the penalty level 
$\lam$ in Theorems \ref{th-adaptive-Lasso} and \ref{th-recursive-Lasso} are
comparable with the best proven results and of the smallest possible order
in linear regression. For $\log(p/n)\ll \log p$, the proper penalty level is
expected to be of the order $\sigma\sqrt{(2/n)\log(p/|S_0|)}$ under a vectorized
sub-Gaussian condition which is slightly stronger than (\ref{subGaussian}). This
refinement for smaller $p$ is beyond the scope of this paper.
\end{remark}

\begin{remark}
The constant factors used in Theorems \ref{th-adaptive-Lasso} and \ref{th-recursive-Lasso}
provide conditions of slightly weaker form than those based on sparse eigenvalues,
although they typically do no imply each other due to differences in the dimension
of covered models and various constant factors.
If $\phi_0(b)=M_3|b_S|_1$
can be used as in (\ref{phi_0-2}), then $M_3|S|F_0(\xi,S;\phi_0)\ge F_2(\xi,S)$.
In GLM, $\phi_0=M_2|b|_2$ can be used as in (\ref{GIF-2}) to weaken this regularity condition.
Since $|b_S|_1\le |S|^{1/2}|b_S|_2$ and $S_0\subset S$,
$F_0(\xi,S_0;\phi_2)\ge F_0(\xi,S;\phi_2)\ge F_2(\xi,S)$.
\end{remark}

\begin{remark}
Although Theorem \ref{th-adaptive-Lasso} is valid for the smaller
$\xi\ge (A+1-\kappa\gamma_0)/(A-1)$, the proof of Theorem \ref{th-recursive-Lasso}
requires $\xi\ge (A+1)/(A-1)$.
\end{remark}

\section{Selection consistency}
\label{selection-consistency}
In this section, we provide a selection consistency theorem for the $\ell_1$ penalized
convex minimization estimator, including both the weighted and unweighted cases.
Let $\|M\|_\infty = \max_{|u|_\infty\le 1}|Mu|_\infty$ for matrices $M$.

\begin{theorem}\label{th-select}
Let $\hbeta$ be as in (\ref{Lasso}), $\beta^*$ be a target vector,
$z^*_k$ be as in (\ref{z^*}), $\Omega_0$ in (\ref{Omega_0}),
$S=\{j:\beta_j^*\neq 0\}$ and $F(\xi,S;\phi_0,\phi)$ as in (\ref{GIF}). \\
(i) Let $0<\eta\le \eta^*\le 1$ and
$\scrB_0^*=\{\beta: \phi_0(\beta-\beta^*) \le \eta\}$. Suppose
\bel{irrepresB}
&& \sup_{\beta\in\scrB_0^*} \big\|W_{S^c}^{-1}\ddpsi_{S^c,S}(\beta)
\{\ddpsi_{{S}}(\beta)\}^{-1}W_S\big\|_\infty \le \kappa_0 < 1,
\\ \label{select-cond-outB}
&& \sup_{\beta\in\scrB_0^*} \big\|W_{S^c}^{-1}\ddpsi_{S^c,S}(\beta)
\{\ddpsi_{{S}}(\beta)\}^{-1}\big\|_\infty\le \kappa_1.
\eel
Then, $\{j:\hbeta_j\neq 0\}\subseteq S$ in the event
\bel{select-th-2}
\Omega_1^* = \Omega_0\cap \Big\{|w_S|_\infty\lam+z^*_0 \le
\eta e^{-\eta}F(0, S;\phi_0,\phi_0),\
\kappa_1z^*_0 + z^*_1 \le (1-\kappa_0)\lam\Big\}.
\eel
(ii) Let $0<\eta\le \eta^*\le 1$ and
$\scrB_0=\{\beta: \phi_0(\beta-\beta^*) \le \eta, \sgn(\beta)=\sgn(\beta^*)\}$.
Suppose (\ref{irrepresB}) and (\ref{select-cond-outB}) hold with $\scrB_0^*$
replaced by $\scrB_0$ and
\bel{select-cond-in}
&& \sup_{\beta\in\scrB_0}\big\|\{\ddpsi_{{S}}(\beta)\}^{-1}\big\|_\infty \le M_0,
\eel
Then, $\sgn(\hbeta)=\sgn(\beta^*)$ in the event
\bel{select-th-1}
\Omega_1^*\cap
\big\{|w_S|_\infty\lam+z^*_0 < M_0^{-1}\min_{j\in S}|\beta^*_j|\big\}.
\eel
(iii) Suppose conditions of Theorem \ref{th-est-GLM} hold for the GLM.
Then, the conclusions of (i) and (ii) hold under the respective conditions if
$F(0,S;\phi_0,\phi_0)$ is replaced by $F^*(\xi,S;M_2|\cdot|_2)$ or
$F_*(\xi,S)$ or $\kappa_*^2(\xi,S)/(M_3|S|)$ with the respective
$\phi_0$ in Theorem \ref{th-est-GLM}.
\end{theorem}

For $w_j=1$, this result is somewhat more specific in the radius $\eta$
for the uniforn irrepresentable conditon (\ref{irrepresB}), compared with
a similar extension of the selection consistency theory to the graphical
Lasso by \cite{RavikumarWRY08}.
In linear regression (\ref{linear-reg}), $\ddpsi(\beta)=\Sigma=X'X/n$ does not
depend on $\beta$, so that Theorem \ref{th-select} with the special $w_j=1$
matches the existing selection consistency theory for the unweighted
Lasso \cite{MeinshausenB06, Tropp06, ZhaoY06,Wainwright09}.
We discuss below the $\ell_1$ penalized logistic regression
as a specific example.

\begin{example}
\label{logistic-sel-consist}
\textbf{(Logistic regression: selection consistency)}
Suppose $w_j=1=|x_j|_2^2/n$ where $x_j$ are the columns of $X$.
If (\ref{select-th-2}) and (\ref{select-th-1}) hold with $z^*_0$ and $z^*_1$
replaced by $\sqrt{(\log(p/\eps_0))/(2n)}$, then the respective conclusions
of Theorem \ref{th-select} hold with at least probability $1-\eps_0$ in
$P_{\beta^*}$.
\end{example}

\section{The sparsity of the Lasso and SRC}
\label{sparsity}
The results in Sections 2 and 3 are concerned with the
estimation and prediction properties of $\hbeta$, but not
dimension reduction. In this section, we provide
upper bound for the dimension of $\hbeta$.
For this purpose, we need to strengthen (\ref{Hessian-cond-1}) to
\bel{Hessian-cond}
e^{-\phi_0(b)}\Sigma^* \le \ddpsi(\beta^*+b) \le e^{\phi_0(b)}\Sigma^*,\quad
\forall\ b\in\scrC(\xi,S),\ \phi_0(b) \le \eta^*.
\eel
We assume the following sparse Riesz condition, or SRC
\cite{ZhangH08}:
\bel{SRC}
c_*\le u'\ddpsi_A(\beta^*)u \le c^*,\
\frac{|S|}{2(1-\alpha)}\Big(\frac{e^{2\eta}c^*}{c_*}+1-\alpha\Big)\le d^*
\eel
for certain constants $\{c_*,c^*\}$, integer $d^*$, $0<\alpha\le 1$, $0<\eta\le\eta^*\le 1$,
all $A\supset S$ with $|A|=d^*$ and all $u\in\real^A$ with $|u|=1$.
The following theorem is an extension of the dimension bounds in \cite{Zhang10-mc+}
from linear regression.

\begin{theorem}\label{sparse1}
Let $\beta^*$ and $S$ be as in Theorem \ref{th-est}.
Consider the $\hbeta$ defined in
(\ref{Lasso}) with $w_j=1$ for all $j$.
Suppose (\ref{Hessian-cond}) and (\ref{SRC}) hold. Then,
\bes
\#\{j: \hbeta_j\neq 0, j\not\in S\} \le d_1
= \Big\lfloor \frac{|S|}{2(1-\alpha)}\Big(\frac{e^{2\eta}c^*}{c_*}-1\Big)\Big\rfloor
\ees
in the event $\Omega_1$ is defined in (\ref{noise-cond}), provided that
\bes
\max_{A\supset S, |A|\le d_1} |(\Sigma^*)_A^{-1/2}\dell_A(\beta^*)|_2
\le e^{-\eta} \alpha\lam\sqrt{(d_1-|S|)/c^*}. 
\ees
\end{theorem}

For GLM, the results on the dimension bounds of the Lasso can be slightly
simplified. Let $\lam_{\xi}=(\xi-1)\lam/(\xi+1)$.  Suppose (\ref{Hessian-cond}) and (\ref{SRC}) hold and
$(\lam+\lam_{\xi})   
\le M_1\eta e^{-\eta} F_*(0, S)$
with $0 < \eta \le 1$. Then,
\bes
\#\{j: \hbeta_j\neq 0, j\not\in S\} \le d_1
= \Big\lfloor \frac{|S|}{2(1-\alpha)}\Big(\frac{e^{2\eta}c^*}{c_*}-1\Big)\Big\rfloor
\ees
in the event $\{z^* < \lam_{\xi}\}$.
The probability of the event $\{z^* < \lam_{\xi}\}$
can be calculated using Lemma \ref{Lem2} as in the previous sections.

\section{Discussion}
\label{discussion}

In this paper, we studied the estimation, prediction, selection and
sparsity properties of  the weighted $\ell_1$-penalized estimators in a general
convex loss formulation.

We applied our general results to several important
statistical models, including linear regression and generalized linear
models. For linear regression, we extend the existing results to weighted/adaptive Lasso.
For the GLMs, the $\ell_q, q \ge 1$ error bounds for a general $q \ge 1$ for the GLMs are not available in the literature, although $\ell_1$ and $\ell_2$ bounds
have been obtained under different sets of conditions respectively in
\cite{vandeGeer08, NegahbanRWY10}. Our fixed-sample analysis
provides explicit constant factors in an explicit neighborhood of a target.
Our oracle inequalities yields even sharper results for multistage recursive
application of an adaptive Lasso.

An interesting aspect of the approach taken in this paper in dealing with
general convex losses such as those for the GLM is that the conditions imposed
on the Hessian naturally `converge' to those for the linear regression as
the convex loss `converges' to a quadratic form.

A key quantity used in the derivation of the results is the generalized
invertibility factor (\ref{GIF}), which grow out of the idea of the $\ell_2$
restricted eigenvalue but improves upon it.
The use of GIF yields sharper bounds on the estimation and prediction errors.
This was discussed in detail in the context of linear regression in \cite{vandeGeerB09,YeZ10}.

We assume that the convex function $\psi(\cdot)$ is twice differentiable. Although this assumption is satisfied in many important and widely used statistical models, it would be interesting to extend the results obtained
in this paper to models with  less smooth loss functions, such as those
in quantile regression and support vector machine.

\section{Appendix}

\noindent
{\bf Proof of Lemma \ref{Lem1}.}
Since $\dpsi(\hbeta)-\dpsi(\beta^*)=z-\dpsi(\beta^*)-g$, (\ref{KKT}) implies
\bes
\Delta(\hbeta,\beta^*)
= \big\langle \hbeta, z-\dpsi(\beta^*)\big\rangle - \lam |\hW\hbeta|_1
 - \big\langle \beta^*, z-\dpsi(\beta^*)-g\big\rangle
\ees
and $|g_j|\le \hw_j\lam$. Thus, (\ref{pred}) follows from
$|(z-\dpsi(\beta^*)_j|\le \hw_j\lam$ and $\hw_j\le w_j$ in $S$ in $\Omega_0$.

For (\ref{basic-ineq}), we have $h_{S^c}=\hbeta_{S^c}$ and $\beta^*_{S^c}=0$,
so that in $\Omega_0$ (\ref{KKT}) gives
\bes
\Delta(\hbeta,\beta^*)
&=& \big\langle \hbeta_{S^c}, \{z-\dpsi(\beta^*)\}_{S^c} \big\rangle
- \lam |\hW_{S^c}\hbeta_{S^c}|_1 - \big\langle h_S, \{z-\dpsi(\beta^*)-g\}_S \big\rangle
\cr &\le& |W_{S^c}\hbeta_{S^c}|_1(z_1^*-\lam) + \big\langle h_S, g_S - \{z-\dpsi(\beta^*)\}_S \big\rangle
\cr &\le& |W_{S^c}\hbeta_{S^c}|_1(z_1^*-\lam)+|h_S|_1(z^*_0 + |w_S|_\infty\lam).
\ees
This gives (\ref{basic-ineq}).
Since $\Delta(\hbeta,\beta^*)>0$, $h\in \scrC(\xi,S)$
when $(|w_S|_\infty\lam + z^*_0)/(\lam - z^*_1)\le\xi$.
For $j\not\in S$, $h_j(\dpsi(\beta+h)-\dpsi(\beta))_j=\hbeta_j(z-\dpsi(\beta^*)-g)_j
\le |\hbeta_j|(w_j\lam-g_j)\le 0$. $\hfill\Box$

\medskip\noindent
\textbf{Proof of Theorem \ref{th-est}.} Let $h=\hbeta-\beta^*$.
Since $\psi(\beta)$ is a convex function,
\bes
t^{-1}\Delta(\beta^*+th,\beta^*)
=\frac{\pa}{\pa t}\Big\{\psi(\beta^*+th)
-t\big\langle h,\dpsi(\beta^*)\big\rangle\Big\}
\ees
is an increasing function of $t$. For $0\le t\le 1$ and in the event $\Omega_1$,
(\ref{basic-ineq}) implies
\bes
t^{-1}\Delta(\beta^*+th,\beta^*)
\le \Delta(h+\beta^*,\beta^*) < (|w_S|_\infty\lam + z^*_0)|h_S|_1.
\ees
By (\ref{cone}) and (\ref{GIF}), $F(\xi,S; \phi_0,\phi_0)
\le \Delta(\beta^*+th,\beta^*) e^{\phi_0(th)}/\{t|h_S|_1\phi_0(th)\}$ for $\phi_0(th)\le\eta^*$.
Thus, for $\phi_0(th) \le \min\{\eta^*, \phi_0(h)\}$ and in the event $\Omega_1$,
\bes
\phi_0(th)e^{- \phi_0(th)}
\le \frac{\Delta(\beta^*+th,\beta^*)}{t|h_S|_1 F(\xi,S; \phi_0,\phi_0)}
< \frac{|w_S|_\infty\lam + z^*_0}{F(\xi,S; \phi_0,\phi_0)}
\le \eta e^{-\eta}.
\ees
If $\eta^* < \phi_0(h)$, the above inequality at $\phi_0(th)=\eta^*$
would give $\eta^* e^{-\eta^*} < \eta e^{-\eta}$, which contradicts to
$\eta\le\eta^*\le 1$. Thus, $\eta^* \ge \phi_0(h)$ and
$\phi_0(th)e^{- \phi_0(th)} \le \eta e^{-\eta}$ for all $0\le t\le 1$.
This implies $\phi_0(h)\le\eta\le\eta^*$.
Another application of (\ref{basic-ineq}) yields
\bes
\phi(h)\le \frac{\Delta(\beta^*+h,\beta^*)e^{\phi_0(h)}}
{F(\xi,S;\phi_0,\phi)|h_S|_1}
\le \frac{(|w_S|_\infty\lam + z^*_0)e^\eta}
{F(\xi,S;\phi_0,\phi)}.
\ees
We obtain (\ref{oracle-1b}) by applying (\ref{oracle-1a}) with $\phi=\phi_{1,S}$
to the right-hand side of (\ref{basic-ineq}). $\hfill\Box$

\medskip\noindent
{\bf Proof of Lemma \ref{Lem2}.}
(i) Since $\dpsi(\beta)=\sum_{i=1}^n x^i\dpsi_0(x^i\beta)/n$ by (\ref{GLM-1}),
\bel{pf-lm-2-1}
\rE_{\beta}\exp\Big\{\frac{n}{\sigma^2}b'(z-\dpsi(\beta))\Big\}
&=& \exp\Big[\sum_{i=1}^n \frac{\psi_0(x^i(\beta+b))
- \psi_0(x^i\beta) - (x^ib)\dpsi_0(x^i\beta)}{\sigma^2}\Big]
\cr &=& \exp\Big[\sum_{i=1}^n \int_0^1\frac{(x^ib)^2\ddpsi_0(x^i(\beta+tb))}{\sigma^2}(1-t)dt\Big].
\eel
This and (\ref{cond-6}) imply that for $M_1|Xb|_\infty\le \eta_0$,
\bel{pf-lm-2-2}
\rE_{\beta^*}\exp\Big\{\frac{n}{\sigma^2}b'(z-\dpsi(\beta^*))\Big\}
\le \exp\Big[\frac{ne^{\eta_0}\langle b, \Sigma^* b\rangle }{2\sigma^2}\Big].
\eel
Since $\max_{k=0,1}z^*_k/\lam_k = \max_j t_j^{-1}|z_j - \dpsi_j(\beta^*)|$ by (\ref{z^*}),
\bes
\rP_{\beta^*}\Big\{ \max_{k=0,1}z^*_k/\lam_k > 1\Big\}
&\le& \sum_{j=1}^p \rP_{\beta^*}\Big\{ |z_j-\dpsi_j(\beta^*)| > t_j \Big\}
\cr &\le& \sum_{j=1}^p \rE_{\beta^*}
\exp\Big\{\frac{n}{\sigma^2}b_j |z_j-\dpsi_j(\beta^*)| - \frac{n}{\sigma^2}b_jt_j\Big\}
\ees
with $b_j = e^{- \eta_0}t_j/\Sigma^*_{jj}$.
Since $M_1\max_{ij}|x_{ij}|b_j\le \eta_0$, (\ref{pf-lm-2-2}) gives
\bes
\rP_{\beta^*}\Big\{ \max_{k=0,1}z^*_k/\lam_k > 1\Big\}
\le \sum_{j=1}^p 2\exp\Big( - \frac{ne^{- \eta_0}t_j^2}{2\sigma^2 \Sigma^*_{jj}}\Big).
\ees
(ii) If (\ref{lm-2-cond-2}) holds, we simply replace $\ddpsi_0(x^i(\beta+tb))$ by $c_0$
in (\ref{pf-lm-2-1}). The rest is simpler and omitted.  $\hfill\Box$

\medskip\noindent
{\bf Proof of Theorem \ref{th-est-GLM}.} (i) Since $F^*(\xi,S;\phi)$ in (\ref{GIF-2})
is a lower bound of $F(\xi,S;\phi_0,\phi)$ in (\ref{GIF}),
(\ref{th-est-GLM-2}) follows from Theorem \ref{th-est} with $\phi_0(b)=M_2|b|_2$.
The probability statement follows from Lemma \ref{Lem2}.
(ii) Since (\ref{Hessian-cond-1}) holds for
the $\phi_0(b)$ in (\ref{phi_0-1}), we are allowed to use $F_*(\xi,S)=F_0(\xi,S;\phi_0)$
in Corollary \ref{Cor1}.
The condition ${\eta^*}=\infty$ is used since $\phi_0(b)$ does not control
$M_1|Xb|_\infty$.
(iii) We are also allowed to use $\phi_0(b)=M_3|b_S|_1$ in
(\ref{phi_0-2}) due to $M_1|Xb|_\infty\le\phi_0(b)$.
$\hfill\Box$

\medskip\noindent
\textbf{Proof of Theorem \ref{th-adaptive-Lasso}.}
Let $h=\hbeta - \beta^*$, $w_j=\what_j$ and $S = \{j: |\hbeta_j|>\gamma_0 \lam\}\cup S_0$.
For $j\not\in S$, $w_j\lam = \drho_\lam(\tbeta_j) \ge \drho_\lam(0+)-\kappa\gamma_0\lam
=(1-\kappa\gamma_0)\lam$, so that
$z_1^* = |W_{S^c}^{-1}\{z-\dpsi(\beta^*)\}_{S^c}|_\infty\le \lam_0/(1-\kappa\gamma_0) = \lam/A$.
We also have $z_0^*\le (1-\kappa\gamma_0)\lam/A$. Since
$|\what|_\infty\le 1$, these bounds for $z^*_0$ and $z^*_1$ yield
\bes
\frac{|\what_S|_\infty\lam+z_0^*}{\lam - z^*_1}
\le \frac{\lam + (1-\kappa\gamma_0)\lam/A}{\lam - \lam/A}
= \frac{A+1-\kappa\gamma_0}{A-1}\le\xi.
\ees
Thus, by Lemma \ref{Lem1}
\bes
h\in \scrC(\xi,S),\ \Delta(\beta^*+h,\beta^*)
\le |h_S|_2\big(|\what_S|_2\lam+|\{z-\dpsi(\beta^*)\}_S|_2\big)
\ees
Since $|S\setminus S_0| \le |(\tbeta-\beta^*)_{S_0^c}|_2^2/\gamma_0^2\lam^2\le \ell^*$,
we have
\bes
|w_S|_\infty\lam + z^*_0\le \lam+(1-\kappa\gamma_0)\lam/A
\le F_0(\xi,S;\phi_0)\eta e^{-\eta}.
\ees
Thus, $\phi_0(h)\le\eta$ by (\ref{oracle-2}). It follows that
$\Delta(\hbeta,\beta^*)\ge e^{-\eta}h'\Sigma h$ by (\ref{Hessian-cond-1}),
so that by (\ref{F_2}),
\bes
e^{-\eta} F_* |h|_2\le
e^{-\eta} F_2(\xi,S) |h|_2\le h'\Sigma h e^{-\eta}/|h_S|_2
\le \Delta(\beta^*+h,\beta^*)/|h_S|_2
\ees
when $|h_S|\neq 0$.
Consequently,
\bel{pf-th-adaptive-Lasso-1}
e^{-\eta} F_* |h|_2 \le |\what_S|_2\lam+|\{z-\dpsi(\beta^*)\}_S|_2.
\eel

Since $\what_j\lam = \drho_\lam(|\tbeta_j|) \le \drho_\lam(|\beta^*_j|)
+ \kappa|\tbeta_j-\beta^*_j|$, we have
\bes
|\what_S|_2\lam \le |\drho_\lam(|\beta^*_{S_0}|)|_2+\kappa |\tbeta-\beta^*|_2.
\ees
Since $|z-\dpsi(\beta^*)|_\infty\le (1-\kappa\gamma_0)\lam/A$,
\bes
|\{z-\dpsi(\beta^*)\}_S|_2
&\le& |\{z-\dpsi(\beta^*)\}_{S_0}|_2+|S\setminus S_0|^{1/2}(1-\kappa\gamma_0)\lam/A
\cr &\le& |\{z-\dpsi(\beta^*)\}_{S_0}|_2+|\tbeta-\beta^*|_2(1-\kappa\gamma_0)/(\gamma_0A).
\ees
Inserting the above inequalities into (\ref{pf-th-adaptive-Lasso-1}), we find that
\bes
e^{-\eta} F_* |\hbeta-\beta^*|_2\le
|\drho_\lam(|\beta^*_{S_0}|)|_2 + |\{z-\dpsi(\beta^*)\}_{S_0}|_2+
\Big(\kappa+\frac{1}{\gamma_0A}-\frac{\kappa}{A}\Big) |\tbeta-\beta^*|_2.
\ees
The probability statement follows directly from (\ref{subGaussian}) with the
union bound. $\hfill\square$

\medskip\noindent
\textbf{Proof of Theorem \ref{th-recursive-Lasso}.} Let $R^{(\ell)}$ be as
in Remark \ref{th-recursive-Lasso-remark}.
For $|z-\dpsi(\beta^*)|_\infty\le\lam_0$, Corollary \ref{Cor1} gives
\bes
|\hbeta^{(0)}-\beta^*|_2 \le e^\eta(\lam+\lam_0)|S_0|^{1/2}/F_0(\xi,S_0;\phi_2)=R^{(0)}.
\ees
Under conditions (\ref{th-recursive-Lasso-cond-1})
and (\ref{th-recursive-Lasso-event}), we have
$R^{(\ell)}\le  \gamma_0\lam\sqrt{\ell^*}$ for all $\ell\ge 0$.
We prove (\ref{th-recursive-Lasso-1}) by induction. We have already proved
 (\ref{th-recursive-Lasso-1}) for $\ell=0$.
 For $\ell\ge 1$, we let $\tbeta=\hbeta^{(\ell-1)}$
 and apply Theorem \ref{th-adaptive-Lasso}:
 $|\hbeta^{(\ell)}-\beta^*|_2\le (1-r_0)R^{(\infty)} + r_0R^{(\ell-1)}=R^{(\ell)}$.
The probability statement follows directly from (\ref{subGaussian}) with the
union bound. $\hfill\square$

\medskip\noindent
\textbf{Proof of Theorem \ref{th-select}.} We first prove the more complicated part (ii).
Let $\ztil = z - \dpsi(\beta^*)$ and $\lam$ be fixed. Consider
\bel{t-path}
\hbeta(\lam,t) = \argmin_\beta
\Big\{\psi(\beta)- \langle \beta, \dpsi(\beta^*) + t\ztil \rangle
+ t\lam\sum_{j=1}^p \hw_j|\beta_j|: \beta_{S^c}=0\Big\}
\eel
as an artificial path for $0\le t\le 1$.
For each $t$, the KKT conditions for $\hbeta(\lam,t)$ are
\bes
g_j(\lam,t) \begin{cases} = t \hw_j\lam \sgn(\hbeta_j(\lam,t)) & \forall \hbeta_j(\lam,t)\neq 0
\cr \in t\hw_j[-1,1], & \forall j\in S, \end{cases}
\ees
where $g(\lam,t) = - \dpsi(\hbeta(\lam,t)) + \dpsi(\beta^*) + t\ztil$.
Let $h(\lam,t)=\hbeta(\lam,t)-\beta^*$.
Since $h_{S^c}=0$, the proof of Theorem \ref{th-est} for $\xi=0$ yields
\bel{pf-select-1}
\phi_0(\hbeta(\lam,t)-\beta^*) \le \eta,\ \forall 0<t \le 1.
\eel

Since $\ddpsi_S(\beta^*)$ is positive-definite, $\hbeta(\lam,0+)=\beta^*$.
It follows that $\sgn(\hbeta_S(\lam,t))=\sgn(\beta^*_S)$ for $0<t < t_1$ for a certain $0<t_1\le 1$.
An application of the differentiation operator $D=(\pa/\pa t)$ to the KKT condition yields
\bes
\ztil_j - \ddpsi_{j,S}(\hbeta(\lam,t))\{(D\hbeta)(\lam,t)\}_S
= \hw_j\lam\, \sgn(\beta^*_j),\ \forall j\in S, 0<t < t_1.
\ees
Thus, for $0<t < t_1$
\bel{pf-select-2}
(D\hbeta)_S(\lam,t)
= \{\ddpsi_{{S}}(\hbeta(\lam,t))\}^{-1}\{\ztil_S - \lam \hW_S\,\sgn(\beta^*_S)\}
\eel
and with an application of the chain rule,
\bel{pf-select-3}
D\dell_{S^c}(\hbeta(\lam,t))
&=& \ddell_{S^c,S}(\hbeta(\lam,t))(D\hbeta)_S(\lam,t)
\cr &=&\ddpsi_{S^c,S}(\hbeta(\lam,t))\{\ddpsi_{{S}}(\hbeta(\lam,t))\}^{-1}
\{\ztil_S - \lam \hW_S\,\sgn(\beta^*_S)\}.
\eel
By (\ref{pf-select-1}), $\hbeta(\lam,t)\in\scrB_0$ for $0<t<t_1$.
It follows from (\ref{pf-select-2}), (\ref{select-cond-in}) and (\ref{select-th-1}) that
\bes
|(D\hbeta)_S(\lam,t)|_\infty\le M_0|\ztil_S - \lam \hW_S\,\sgn(\beta^*_S)|_\infty
\le M_0(|\hw_S|_\infty\lam+z^*_0) < \min_{j\in S}|\beta^*_j| - \eps_1
\ees
for $0<t < t_1$ and some $\eps_1>0$. Thus,
$|h_S(\lam,t)|_\infty\le t M_0(|w_S|_\infty\lam+z^*_0) < \min_{j\in S}|\beta^*_j|-\eps_1$.
This implies $\sgn(\hbeta(\lam,t-))=\sgn(\beta^*)$ for $0 < t \le 1$
by the continuity of $\hbeta(\lam,t)$ in $t$, i.e.\ $ t_1=1$.
Since $|W_S^{-1}\hW_Sv_S|_\infty\le |v_S|_\infty$ for all $v\in\real^p$ in $\Omega_0$,
(\ref{pf-select-3}), (\ref{irrepresB}) and (\ref{select-cond-outB}) implies
that for $0<t < 1$
\bes
|W_{S^c}^{-1}D\dell_{S^c}(\hbeta(\lam,t))|
&\le & |W_{S^c}^{-1}\ddpsi_{S^c,S}(\hbeta(\lam,t))\{\ddpsi_{{S}}(\hbeta(\lam,t))\}^{-1}\ztil_S|_\infty
\cr && + \lam |W_{S^c}^{-1}\ddpsi_{S^c,S}(\hbeta(\lam,t))
\{\ddpsi_{{S}}(\hbeta(\lam,t))\}^{-1}W_S\,\sgn(\beta^*_S)|_\infty
\cr &\le& \kappa_1z^*_0 + \kappa_0\lam.
\ees
This implies $|W_{S^c}^{-1}\dell_{S^c}(\hbeta(\lam,1))|_\infty
\le \kappa_1z^*_0+\kappa_0\lam+|W_{S^c}^{-1}\dell_{S^c}(\beta^*)|_\infty
\le \kappa_1z^*_0+z^*_1+\kappa_0\lam\le \lam$. It follows that
\bes
\begin{cases}
\dell_j(\hbeta(\lam,1-)) = \hw_j\lam\sgn(\beta^*_j), \
\sgn(\beta^*_j)=\sgn(\hbeta(\lam,1-)), &  j\in S
\cr
\dell_j(\hbeta(\lam,1-)) \in \hw_j\lam[-1,1], & j\not\in S.
\end{cases}
\ees
These are the KKT conditions for $\hbeta(\lam,1-)$ with $\sgn(\hbeta(\lam,1-))=\sgn(\beta^*)$.

The proof for part (ii) is similar, with $\sgn(\beta^*)$ replaced by $\sgn(\hbeta(\lam,t))$ in
the proof of part (i).  Finally, in part (iii), $F_0(\xi,S;\phi_0,\phi_0)$ is simply
replaced by its lower bounds with the respective $\phi_0$. $\hfill\square$

\medskip
\noindent
\textbf{Proof of Theorem \ref{sparse1}.}  Let $A_1=\{j: |g_j|=\lam\}\cup S$, $A_0=A_1\setminus S$ and
$\Sigmahat = \int_0^1\ddpsi(\beta^*+t h)dt$, where $g$ is the negative gradient in (\ref{KKT})
and $h=\hbeta-\beta^*$. Let $g_{(A)}=(g_jI\{j\in A\})'$.
Consider the case $|A_1|\le d^*$ (e.g. with sufficiently small $z^*$).
Since $g_{A_0}h_{A_0}=\lam| h_{A_0}|_1$ by (\ref{KKT})
and $\{\dell(\beta^*+h)-\dell(\beta^*)\}_{A_1}=\Sigmahat_{A_1} h_{A_1}$,
\bes
g_{(A_0)}\Sigmahat_{A_1}^{-1}g_{(A_1)}
= - g_{(A_0)}\Sigmahat_{A_1}^{-1}\dell_{A_1}(\beta^*+h)
\le - \lam|h_{A_0}|_1+|\Sigmahat_{A_1}^{-1/2}g_{(A_0)}||\Sigmahat_{A_1}^{-1/2}\dell_{A_1}(\beta^*)|.
\ees
Since $|\Sigmahat_{A_1}^{-1/2}g_{(A_0)}|_2^2+\ |\Sigmahat_{A_1}^{-1/2}g_{(A_1)}|_2^2
= |\Sigmahat_{A_1}^{-1/2}g_{(S)}|_2^2+ 2g_{(A_0)}\Sigmahat_{A_1}^{-1}g_{(A_1)}$,
we have
\bes
&& |\Sigmahat_{A_1}^{-1/2}g_{(A_0)}|_2^2 + |\Sigmahat_{A_1}^{-1/2}g_{(A_1)}|_2^2
\le |\Sigmahat_{A_1}^{-1/2}g_{(S)}|_2^2
+ 2|\Sigmahat_{A_1}^{-1/2}g_{(A_0)}||\Sigmahat_{A_1}^{-1/2}\dell_{A_1}(\beta^*)|.
\ees
Thus, in the event $|\Sigmahat_{A_1}^{-1/2}\dell_{A_1}(\beta^*)|
\le \alpha\lam\sqrt{|A_0|/(c^*e^\eta)}$
with $0<\alpha<1$, we have
\bes
(1-\alpha)|\Sigmahat_{A_1}^{-1/2}g_{(A_0)}|_2^2 + |\Sigmahat_{A_1}^{-1/2}g_{(A_1)}|_2^2
\le  |\Sigmahat_{A_1}^{-1/2}g_{(S)}|_2^2 + \alpha \lam^2 |A_0|/(c_*e^\eta).
\ees
Since the eigenvalues of $\Sigmahat_{A_1}$ lie in the interval $c_*e^{-\eta}$ and $c^*e^\eta$
and $g_{A_0}=\lam\,\sgn(\hbeta_{A_0})$,
\bes
\frac{(1-\alpha)\lam^2 |A_0|}{c^*e^\eta}+\frac{\lam^2 |A_0| + |g_{S}|_2^2}{c_* e^\eta}
\le \frac{|g_S|_2^2}{c_*e^{-\eta}}+\frac{\alpha\lam^2 |A_0|}{c^*e^\eta}.
\ees
This gives
\bes
2(1-\alpha)|A_0|\le \Big(\frac{c^*e^{2\eta}}{c_*}-1\Big)\frac{|g_S|_2^2}{\lam^2}
\le \Big(\frac{c^*e^{2\eta}}{c_*}-1\Big)|S|.
\ees
We note that $|\Sigmahat_{A_1}^{-1/2}\dell_{A_1}(\beta^*)|
\le e^{\eta/2}\max_{A\supset S, |A|\le d^*} |(\Sigma^*)_A^{-1/2}\dell_A(\beta^*)|$.
We complete the proof by considering the artificial path (\ref{t-path}). $\hfill\square$

\bibliographystyle{amsalpha}
\bibliography{JMLR-rev}

\end{document}